\documentclass[10pt,twocolumn,letterpaper]{article}

\usepackage{iccv}
\usepackage{times}
\usepackage{epsfig}
\usepackage{graphicx}
\usepackage{amsmath}
\usepackage{amssymb}

\usepackage{microtype}
\usepackage{graphicx}
\usepackage{amsmath}
\usepackage{anyfontsize}
\usepackage{array}
\usepackage{caption}
\usepackage{tabularx}
\usepackage{multirow}
\usepackage{setspace}
\usepackage{amsmath,amssymb}
\usepackage{xcolor}
\usepackage{subfigure}

\usepackage{wrapfig, lipsum}
\usepackage{url, float}
\usepackage{enumitem}
\usepackage{adjustbox}
\usepackage{mathtools}

\definecolor{dartmouthgreen}{rgb}{0.05, 0.5, 0.06}

\newcommand{\fig}[1]{Figure~\ref{fig:#1}}
\newcommand{\sect}[1]{Section~\ref{sect:#1}}
\newcommand{\tab}[1]{Table~\ref{tab:#1}}

\newcommand{\eq}[1]{(\ref{eq:#1})}
\newcommand{\app}[1]{Appendix~\ref{appendix:#1}}

\usepackage[ruled,vlined]{algorithm2e}
\usepackage{booktabs} % for professional tables

\usepackage[pagebackref=true,breaklinks=true,letterpaper=true,colorlinks,bookmarks=false]{hyperref}

\iccvfinalcopy % *** Uncomment this line for the final submission

\def\iccvPaperID{1337} % *** Enter the ICCV Paper ID here

\ificcvfinal\fi

\begin{document}

%%%%%%%%% TITLE
\title{Towards Real-time Text-driven Image Manipulation \\ with Unconditional Diffusion Models}

\author{Nikita Starodubtsev\thanks{Correspondence to {\tt\footnotesize nstarodubtcev@itmo.ru}}\\
Yandex\\
\and
Dmitry Baranchuk\\
Yandex\\
% {\tt\small dbaranchuk@yandex-team.ru}
\and
Valentin Khrulkov\\
Yandex\\
% {\tt\small vkhrulkov@yandex-team.ru}
\and
Artem Babenko\\
Yandex\\
% {\tt\small arbabenko@yandex-team.ru}
}

\maketitle
\ificcvfinal\thispagestyle{empty}\fi

%%%%%%%%% ABSTRACT
\begin{abstract}
Recent advances in diffusion models enable many powerful instruments for image editing. One of these instruments is text-driven image manipulations: editing semantic attributes of an image according to the provided text description. 
% Popular text-conditional diffusion models offer various high-quality image manipulation methods for a broad range of text prompts.
Existing diffusion-based methods already achieve high-quality image manipulations for a broad range of text prompts. 
However, in practice, these methods require high computation costs even with a high-end GPU.
This greatly limits potential real-world applications of diffusion-based image editing, especially when running on user devices.

In this paper, we address efficiency of the recent text-driven editing methods based on unconditional diffusion models and develop a novel algorithm that learns image manipulations $4.5{-}10\times$ faster and applies them $8\times$ faster.
We carefully evaluate the visual quality and expressiveness of our approach on multiple datasets using human annotators.  
Our experiments demonstrate that our algorithm achieves the quality of much more expensive methods. 
Finally, we show that our approach can adapt the pretrained model to the user-specified image and text description on the fly just for ${\sim}4$ seconds.
In this setting, we notice that more compact unconditional diffusion models can be considered as a rational alternative to the popular text-conditional counterparts. 
\end{abstract}

% Recent advances in diffusion models enable many powerful instruments for image editing. One of these instruments is text-driven image manipulations: editing semantic attributes of an image based on the provided text description. Existing diffusion-based methods already achieve high-quality image manipulations for a broad range of text prompts. 
% %The main challenge lies in a necessity of multiple forward passes of a neural network at each step of the fine-tuning.
% %Such requirements lead to the significant computational and time costs and as a consequence make diffusion models less preferable compare to GAN analogues.
% However, in practice, these methods require high computation costs even with a high-end GPU.
% This greatly limits potential real-world applications of diffusion-based image editing, especially when running on user devices.
% In this paper, we develop a novel algorithm that learns image manipulations $4.5{-}10\times$ faster and applies them $8\times$ faster than state-of-the-art.
% We carefully evaluate the visual quality and expressiveness of our approach on a range of datasets using human annotators. 
% Our experiments demonstrate that the proposed algorithm achieves the similar quality as state-of-the-art diffusion-based image manipulations. 
% Finally, we demonstrate that our approach can adapt to the user-specified image and text description on the fly just for ${\sim}4$ seconds.

%%%%%%%%% BODY TEXT
\section{Introduction}
\begin{figure}[ht!]
    \centering
    \includegraphics[width=\columnwidth]{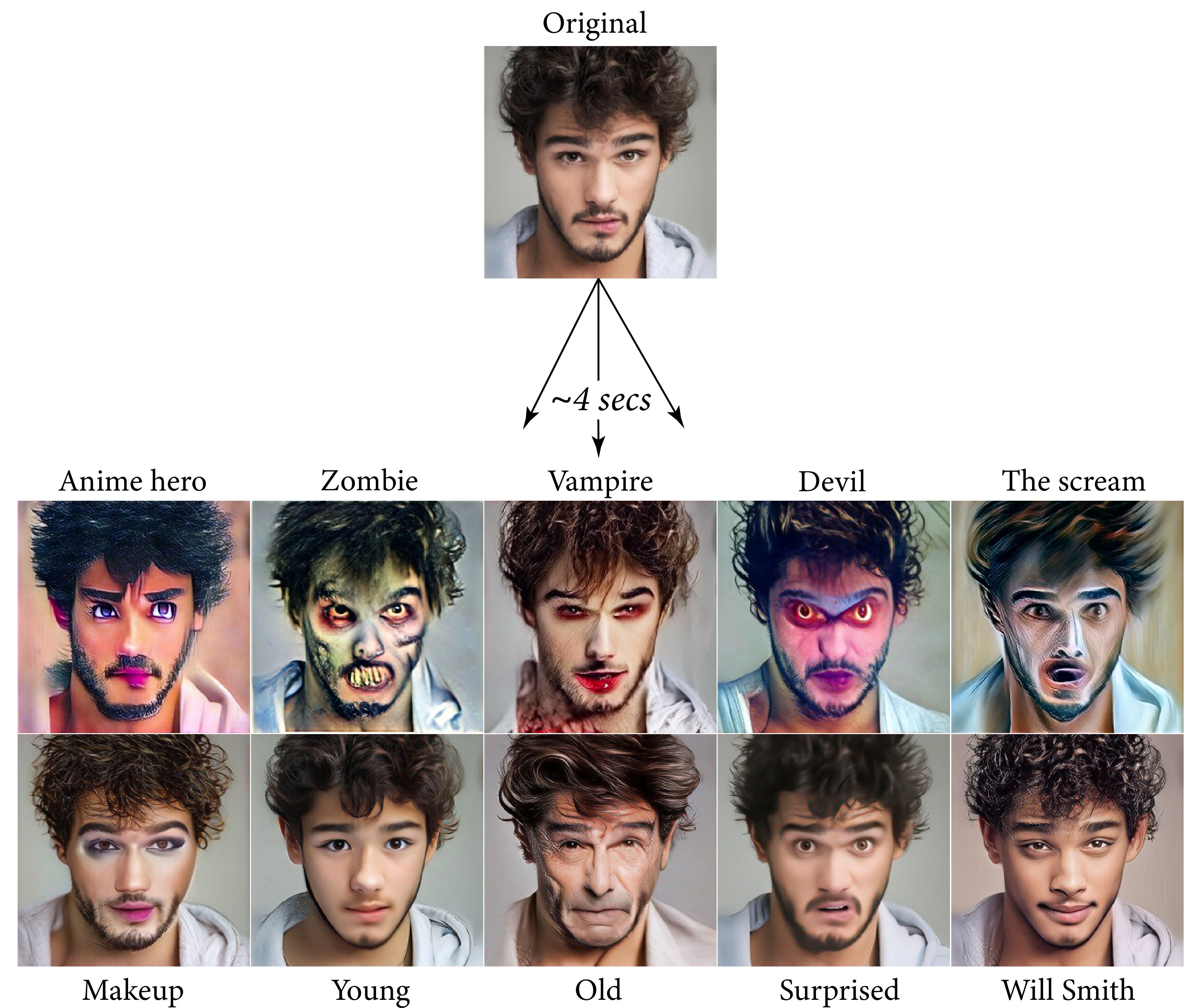}
    \caption{\textbf{Few examples of single image editing.} A user provides an image and a text description of the desired transform. Our diffusion-based approach adapts the pretrained model to the given image and text and returns the manipulated image. The entire procedure takes ${\sim}4$ seconds.}
    \label{fig:my_label}
\end{figure}

Diffusion probabilistic models (DPMs) have recently shown remarkable results in image generation, producing both realistic and diverse samples~\cite{ho2020ddpm,song2021denoising}.
After their initial success, diffusion models have seen use in many related applications, including object detection~\cite{chen2022diffusiondet}, semantic segmentation~\cite{baranchuk2022labelefficient}, image processing~\cite{saharia2021image, lugmayr2022repaint} and many others.

One popular application of diffusion models is text-guided image manipulation, where DPM is used to edit an existing image, changing its attributes to match a user-specified text description.
This application has become a hot topic after the appearance of high-quality text-conditional DPMs like Imagen~\cite{saharia2022photorealistic}, DALL-E 2~\cite{ramesh2022dalle2}, Stable Diffusion~\cite{rombach2022high}.

Text-conditional DPMs are trained on large and diverse datasets with rich text information, such as MS-COCO~\cite{cocodataset} or LAION5B~\cite{schuhmann2022laionb}, and applicable to a wide range of domains at a time.
However, some practical applications focus solely on some simple and narrow domain, e.g., human faces, and often have strict computational budgets.  
In these cases, unconditional diffusion models might be a reasonable alternative since they already provide strong generative performance on such domains being more efficient than large and diverse text-conditional models.
Moreover, unconditional models avoid tedious text annotation process that increases the costs and can raise additional legal and ethical concerns. 

Several recent works have learned text-driven image manipulations using only unconditional generative models.
Some methods use unconditional GANs~\cite{Karras2019stylegan2,Karras2021}, showing strong results for text-driven domain adaptation~\cite{gal2021stylegannada,alanov2022hyperdomainnet,alaluf2022times} and single image editing~\cite{Patashnik_2021_ICCV}.
Other works~\cite{kim2022diffusionclip, kwon2023diffusion} exploit unconditional diffusion models, significantly outperforming GANs in terms of both image fidelity and editing quality. 

However, compared to GANs, diffusion-based image manipulations require expensive sequential inversion and generation, limiting their applicability in resource-limited environments.
Moreover, some applications, e.g., single image editing, also require highly performant online adaptation where current diffusion-based methods are still not efficient enough in terms of both adaptation time and memory usage. 

% However, such powerful models are not always available for the editing task at hand. Text-conditional DPMs require large training datasets with rich text information, such as MS-COCO~\cite{cocodataset} or LAION5B~\cite{schuhmann2022laionb}. In practice, there are many popular image domains where such datasets are unavailable, making it impossible to train text-conditional DPMs. Notable examples of such domains include human faces, medical and satellite imagery.

% To address this issue, several recent works have developed text-driven image manipulations using only fully unconditional generative models.
% unconditional generative models for the moderately complex domains where they already demonstrate high fidelity image generation without any supervision~\cite{}. 
% Therefore, text-driven image manipulation for unconditional generative models is still a high demand topic for real-world applications. 
 
In this work, we analyze the performance bottlenecks of the existing text-guided image manipulation methods based on unconditional diffusion models and propose a workaround to circumvent them. 
As a result, we present a fast and memory-efficient diffusion-based approach that preserves the visual and editing quality of the prior image manipulation methods.

% In this work, we focus on developing a faster diffusion-based image manipulation approach to address real-world use cases. 
% Our goal is to minimize both runtime and memory usage while preserving the visual quality of text-driven diffusion-based image editing methods. 
% To achieve this, we analyze the performance bottlenecks of existing diffusion-based image manipulation methods and propose a workaround to circumvent them. 

To sum up, the contributions of the paper are as follows:
\begin{itemize}[leftmargin=8.5px]
    \item We develop a practical algorithm for text-driven image manipulation based on fast approximate sampling procedures. 
    Our algorithm learns image manipulations $4.5{-}10\times$ faster and applies them $8\times$ faster than previous methods based on unconditional diffusion models. %DiffusionCLIP (the current state-of-the-art). 
    The source code of our algorithm is available online\footnote{\url{https://github.com/quickjkee/eff-diff-edit}}.
    %Propose two simple yet effective ingredients that make text-driven diffusion-based domain adaptation $\sim 1000$ times more efficient compared to the previous state-of-the-art approach.
    
    \item We conduct a thorough human evaluation on multiple standard image datasets to compare the produced image manipulations in terms of visual quality and their correspondence to the text attribute. 
    Our algorithm achieves similar editing quality as significantly more expensive counterparts. 
    
    \item We observe an emergent property of our training procedure. 
    When the model is trained using our protocol, it also learns to improve the perceptual quality of approximate image estimates without being explicitly enforced to do so. 
    We reveal that it is an implicit side-effect of the directional CLIP loss~\cite{gal2021stylegannada}. 
    From a practical standpoint, this phenomenon helps us to accelerate model inference without perceptible loss in image fidelity.

    \item We demonstrate that unconditional diffusion models can learn text-guided image manipulations from a single image. 
    This allows users to create their own image transformations on the fly, using the same image they intend to manipulate. 
    For this application, our approach can edit images with significantly better quality than existing GAN-based alternatives.
    Notably, on narrow domains like human faces, the proposed method can provide reasonable quality-efficiency trade-off compared to the state-of-the-art text-driven editing methods based on Stable Diffusion.

    % \item Finally, we compare our approach to the state-of-the-art methods using text-conditional models. We demonstrate that on simple domains like human faces, the proposed method can provide reasonable quality-efficiency trade-off.
    % Moreover, we show that our method can provide some transforms that the recent text-conditional models fail to produce.
    
    % \item Demonstrate that unconditional diffusion models can learn text-based image manipulation from a single image. 
    % This opens several potential applications that allow users to create their own image manipulations on the fly, using the same image they intend to manipulate. 
\end{itemize}

\section{Background}

This section provides an overview of diffusion models, briefly discusses previous text-driven image manipulation methods and describes the core aspects of the DiffusionCLIP~\cite{kim2022diffusionclip} approach that is mostly leveraged in our work. 

\subsection{Diffusion probabilistic models}

Diffusion models~\cite{ho2020ddpm} are latent variable generative models  trained to approximate the data distribution $x_0{\sim}q(x_0)$ by using \textit{forward} and \textit{reverse} diffusion processes. 

The forward diffusion process $q(x_{1:T}|x_0)$ gradually applies Gaussian noise with some predefined variance schedule $\beta$ to a real image $x_0$ until it converges to isotropic Gaussian distribution $x_T{\sim}N(0, I)$.
A latent variable $x_t$ can be sampled directly from $x_0$ in the closed form:
\begin{equation}
q(x_t|x_0){=}\mathcal{N}(x_t; \sqrt{\bar{\alpha}_t}x_0, 1-\bar{\alpha}_tI),    
\label{eq:ddpm_xt}
\end{equation}
where $\alpha_t \coloneqq 1 - \beta_t$, $\bar{\alpha}_t \coloneqq \prod_{s=1}^t \alpha_s$.

The reverse process allows generating new samples from $q(x_0)$ by gradually transforming $x_T{\sim}N(0, I)$ to $x_0$, which requires $T$ neural network forward passes.
%At each step $t$, a neural network $\theta$ approximates true $p(x_{t-1}|x_t)$ distribution and aims to remove the corresponding noise component.
%Importantly, the reverse process requires $T$ neural network forward passes to generate a single sample.
Often, $T{=}1000$ and hence the generation takes much more time compared with feed-forward generative models, e.g., GANs~\cite{gans, Karras2019stylegan2}. 

Many recent works have addressed this problem~\cite{dockhorn2022genie,Karras2022edm,song2021denoising,song2021scorebased,watson2022learning,lu2022dpmsolver} and demonstrated convincing results within a few dozens of inference steps.
DDIM~\cite{song2021denoising} is one of the most prevalent samplers that  
generates plausible samples for $\tau_{dec} \approx 50{-}100$ steps without re-training the initial model.

In more detail, DDIM obtains a sample $x_{t-1}$ from $x_t$ in the following way:
\begin{equation}
x_0(t; \theta){=}\frac{x_{t}{-}{\sqrt{1{-}\alpha_{t}}}\epsilon_{\theta}(x_{t})}{\sqrt{\alpha_{t}}}
\label{eq:ddim_x0}
\end{equation}
\begin{equation}
x_{t-1}{=}\sqrt{\alpha_{t-1}}{\cdot}x_0(t; \theta){+}\sqrt{1{-}\alpha_{t-1}}\epsilon_{\theta}(x_{t})
\label{eq:ddim}
\end{equation}
where $x_0(t;\theta)$ is a $x_0$ estimate at a time step $t$ and $\epsilon_\theta(x_t)$ is a prediction of the noise component using a pretrained diffusion model with parameters $\theta$.

In addition, DDIM is deterministic and serves as a de-facto standard inversion method for diffusion models with low reconstruction errors.
Opposed to \eq{ddpm_xt}, DDIM has to sequentially apply the pretrained model to map a real image $x_0{\sim}q(x_0)$ into the latent variable $x_t$:
\begin{equation}
x_{t+1}{=}\sqrt{\alpha_{t+1}}{\cdot}x_0(t; \theta){+}\sqrt{1{-}\alpha_{t+1}}\epsilon_{\theta}(x_{t})
\label{eq:forward_ddim}
\end{equation}
Thus, the image ``inversion'' for diffusion models is much more expensive compared with VAEs~\cite{vae} or GANs~\cite{tov2021designing, alaluf2021restyle}.

% The $x_0$ reconstruction is then obtained by the reverse DDIM process \eq{ddim} 

% $x_0(\theta)$ denotes $x_0$ reconstruction that implicitly depends on parameters $\theta$ of the pretrained diffusion model.

% Also, DDIM is deterministic and serves as a de-facto standard inversion method for diffusion models with low reconstruction errors. 
% DDIM sequentially applies the pretrained model to map a real image $x_0$ into the latent variable $x_t$.
% Thus, the image ``inversion'' for diffusion models is much more expensive compared to VAEs~\cite{vae} or GANs~\cite{tov2021designing, alaluf2021restyle}.

% In more detail, at each encoding step, DDIM obtains a sample $x_{t+1}$ from $x_t$ in the following way:
% \begin{equation}
% x_0(t; \theta){=}\frac{x_{t}{-}{\sqrt{1{-}\alpha_{t}}}\epsilon_{\theta}(x_{t})}{\sqrt{\alpha_{t}}}
% \label{eq:ddim_x0}
% \end{equation}
% \begin{equation}
% x_{t+1}{=}\sqrt{\alpha_{t+1}}{\cdot}x_0(t; \theta){+}\sqrt{1{-}\alpha_{t+1}}\epsilon_{\theta}(x_{t})
% \label{eq:forward_ddim}
% \end{equation}
% where $x_0(t;\theta)$ is a $x_0$ estimate at a time step $t$ and $\epsilon_\theta(x_t)$ is a prediction of the noise component using a pretrained diffusion model with parameters $\theta$.

\subsection{Text-guided image manipulation}

One of the leading directions for semantic image manipulation with generative models is to use natural language prompts to guide the image generation toward the desired transformation. 

The first popular group of methods considers pretrained text-conditional diffusion models~\cite{rombach2022high, ramesh2022dalle2, saharia2022photorealistic, pmlr-v162-nichol22a}. 
Some of them leverage the internal knowledge of the existing models by manipulating text embeddings~\cite{gal2022textual, ruiz2022dreambooth} or cross attention weights~\cite{hertz2022prompt, mokady2022nulltext}. % P2P does not work on real images. mokady2022nulltext solves the problem of the poor consistency with real images during ddim inversion (for SD model) and considers mask-based and P2P editing. 
Other works propose to finetune the model parameters on a single image~\cite{zhang2022sine, Valevski2022UniTuneTI} and perform expressive text-driven manipulations upon it. 
The recent work \cite{brooks2022instructpix2pix} finetunes Stable Diffusion on the large set of GPT-3 powered paired examples. 
This allows them to achieve both efficient and high-quality image transforms.

% EDICT: Exact Diffusion Inversion via Coupled Transformations -- propose exact inversion using NF: 2x more expensive than DDIM. Used for conditional inversion in SD. Address the similar problem as Null-text inversion
% \dmitry{Todo: zero-pix2pix,  Unifying Diffusion Models' Latent Space, with Applications to CycleDiffusion and Guidance. Uncovering the Disentanglement Capability in Text-to-Image Diffusion Models}
% UniTune has poor consistency 

Another research direction is to exploit pretrained unconditional models along with an external model for text guidance, typically CLIP~\cite{pmlr-v139-radford21a}.
In general, the methods encode real images into the model latent space and then either modify the latent representations~\cite{kwon2023diffusion, Patashnik_2021_ICCV} or the model parameters directly~\cite{kim2022diffusionclip, gal2021stylegannada} to minimize the CLIP loss. 

Although GAN-based methods~\cite{Patashnik_2021_ICCV, gal2021stylegannada, alanov2022hyperdomainnet, alaluf2022times} can provide efficient adaptation and inference, they still suffer from imperfect image inversion and provide less expressive and natural image transformations compared with the diffusion models.
On the other hand, DDPM-based methods~\cite{kim2022diffusionclip, kwon2023diffusion} have to perform costly sequential forward and reverse processes to edit a single image. 

There is also a line of work that suggests training an image generator from scratch on a single image~\cite{kulikov2022sinddm, bar2022text2live, wang2022sindiffusion} and then applying the CLIP model for the text-driven editing.
The main limitations of such methods are a time consuming adaptation and a scarce set of available transformations, e.g., style transfer and image harmonization.

%\dmitry{todo maybe add localized text-guided editing: blended diffusion (user specified masks + clip guidance, paint by word (gan + clip + user specified mask), diffedit, glide}

\subsection{DiffusionCLIP} 
\label{sect:diffclip}
DiffusionCLIP~\cite{kim2022diffusionclip} is a recent approach for text-driven image manipulation using unconditional diffusion models.
This method adapts $\epsilon_{\theta}$ to a new diffusion model $\epsilon_{\hat{\theta}}$ such that the transformation of a source image $x_0$ into $x_0(\hat{\theta})$ is associated with the text description $y_{tar}$, e.g., ``Makeup face''.
% Specifically, $x_0(\hat{\theta})$ denotes reconstruction of $x_t$ encoding the DDIM latent variable to $x_0$ obtained with the \emph{original model} $\epsilon_{\theta}$, with the new model $\epsilon_{\hat{\theta}}$.

In more detail, it encodes the image $x_0$ into the latent variable $x_{t_0}$ using DDIM in forward direction~\eq{forward_ddim} with the original diffusion model $\epsilon_{\theta}$. 
In practice, the DDIM encoding is applied for some predefined steps $\tau_{enc}\le t_0$ to obtain $x_{t_0}$.
The typical values of $t_0$ are within $[300, 600]$.
Lower $t_0$ is useful for ``shallow'' image manipulations, e.g., adding makeup. 
Higher $t_0$ is required for strong transformations that significantly affect the structure of semantic attributes, e.g., ``Person'' to ``Zombie''. 

Then, at each training iteration, DiffusionCLIP transforms $x_{t_0}$ to $x_0(\theta)$ using the DDIM generation \eq{ddim} for $\tau_{dec}$ steps and backpropagates through the entire decoding process to minimize the following objective: 
\begin{align}
\mathcal{L}(\theta) = \mathcal{L}_{dir}(x_0(\theta), y_{tar};x_0,y_{ref}) + \lambda{\cdot}\mathcal{L}_{id}(x_0(\theta); x_0)
\label{eq:objective}
\end{align}
where $\mathcal{L}_{dir}$ essentially enforces semantic transfer from $y_{ref}$ to $y_{tar}$ and $\mathcal{L}_{id}$ serves to preserve the attributes of the original image $x_0$. 
$\lambda$ is a hyperparameter that controls the regularization strength. 
Note that $\tau_{enc}$ and $\tau_{dec}$ are hyperparameters of the method allowing us to trade reconstruction quality for processing speed; in practice, typical values are $\tau_{enc} \in [40, 200]$ and $\tau_{dec} \in [6, 40]$ \cite{kim2022diffusionclip}.
 
$\mathcal{L}_{dir}$ is a directional CLIP loss~\cite{gal2021stylegannada} which minimizes the cosine distance between cross-domain directions in the CLIP space. 
Specifically, given the CLIP text encoder $E_T$, the text direction is defined as ${\Delta}T{=}E_T(y_{tar}){-}E_T(y_{ref})$, where 
$y_{ref}$ is a text prompt that represents the source domain in general, e.g., ``Person''. 
The direction between the source $x_0$ and transformed $x_0( \theta)$ images is defined similarly using the CLIP image encoder $E_I$: ${\Delta}I{=}E_I(x_0(\theta)){-}E_I(x_0)$. 
The overall $\mathcal{L}_{dir}$ objective is calculated as follows:
\begin{equation}
\mathcal{L}_{dir}= 1 - \frac{\langle{\Delta}T, {\Delta}I\rangle}{\|{\Delta}T\|_2 \|{\Delta}I\|_2}
\label{eq:clip_loss}
\end{equation}
$\mathcal{L}_{id}$ is typically a reconstruction loss in the pixel space: $L_1{=}\|x_0(\theta) - x_0\|_1$. Optionally, it may include additional domain specific losses, e.g., $\mathcal{L}_{face}(x_0(\theta), x_0)$ that attempts to preserve the face identity in the image $x_0$ for the source domain ``Person'' via a pretrained face recognition network.

\textbf{Efficiency.} 
The resulting algorithm performs image manipulations with state-of-the-art quality but requires significantly more compute and memory than the GAN-based counterparts.
When dealing with $256{\times}256$ images, DiffusionCLIP takes minutes to adapt the model to the target transform and a few seconds to apply it to a single image. 

During training, DiffusionCLIP unrolls several diffusion steps and backpropagates through the decoding process at each model update step. 
To allow this backpropagation, it needs to store intermediate model activations from every step, taking up extra GPU memory.
Thus, even with a batch size $1$, it consumes ${\sim}21$GiB GPU VRAM and hence cannot fit most end-user GPUs.
In contrast, the GAN-based competitor~\cite{gal2021stylegannada} consumes only ${\sim}3$GiB.

The authors of DiffusionCLIP suggest a possible way to reduce memory consumption by independently updating the model parameters at each decoding step. This strategy halves the overall GPU usage but doubles the training time due to the sequential model update steps.

During inference, the main performance bottleneck is the DDIM encoding that requires running a forward pass of the diffusion model at each step to apply noise to the original image. 
As a result, the algorithm spends more than half of the total inference time \textit{applying} noise to the image.

% Applying the learned manipulation to a real image also requires multi-step processing, both in forward and reverse phases. DiffusionCLIP uses $\tau_{for}{=}40{-}200$ steps to encode $x_0$ into the latent variable $x_{t_0}$ and $\tau_{gen}{=}40$ steps for the reverse process.
%The forward process consumes at least half of the total inference time.
%Overall, these limitations motivate us to propose a novel text-driven domain adaptation method that preserves DiffusionCLIP editing expressiveness but makes it a way more efficient and accessible in practice.

% \begin{gather}
% \mathcal{L}_{id}(x_0(\hat{\theta}), x_0)= \lambda_{L_1}\|x_0(\hat{\theta}) - x\| + \lambda_{face}\mathcal{L}_{face}(x_0(\hat{\theta}), x),
% \end{gather}
% where $\lambda_{L_1}$ and $\lambda_{face}$ are hyperparameters to control the regularization strength.

\begin{figure}
    \centering
    \includegraphics[width=\columnwidth]{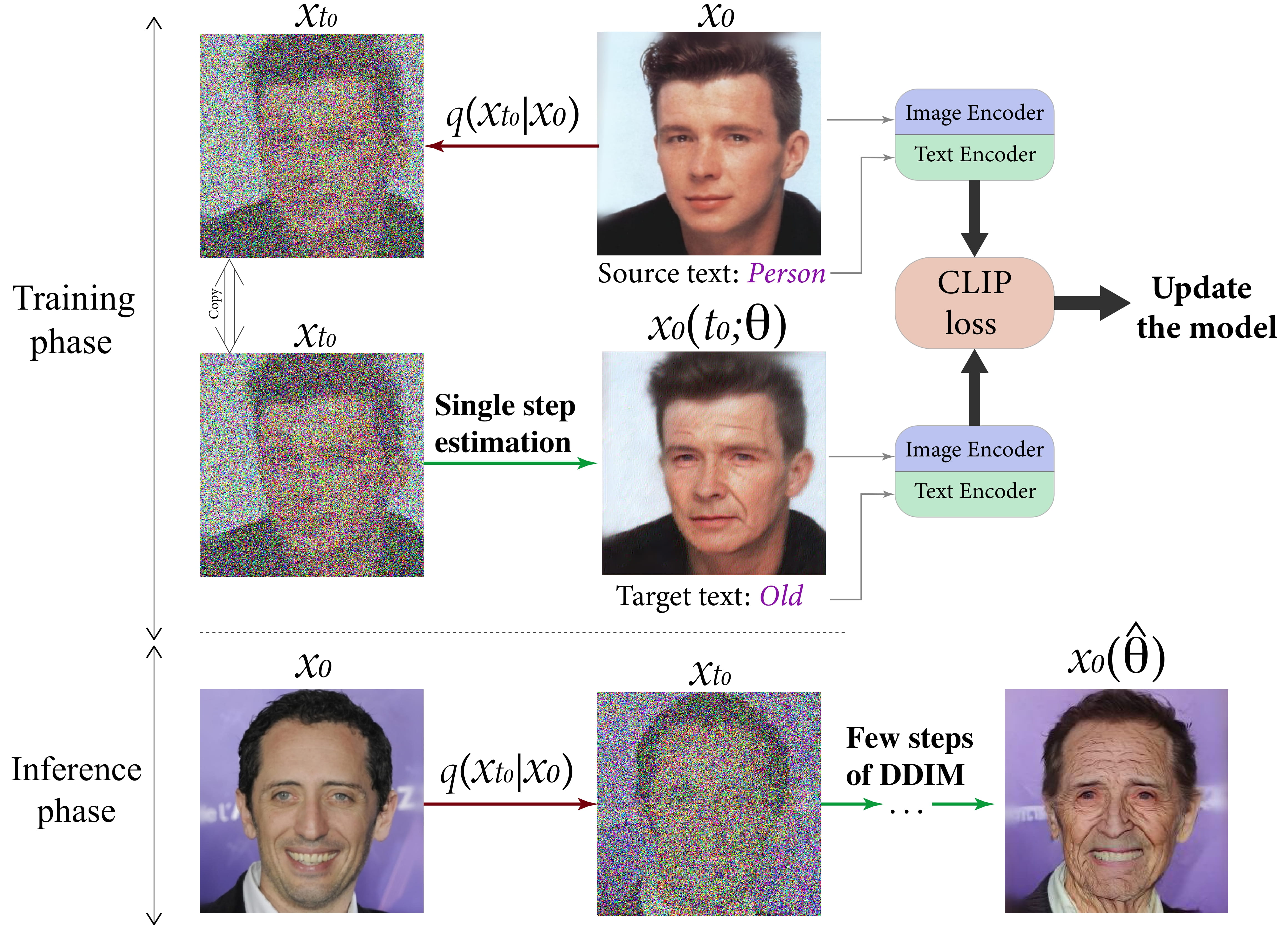}
    \caption{Overview of the proposed approach. \textbf{Training phase.} Diffusion model parameters $\theta$ are updated to minimize the directional CLIP loss~\eq{clip_loss} for a single-step $x_0$ estimate that lack high-frequency perceptual details. \textbf{Inference phase.} A real image $x_0$ is mapped to the latent variable $x_{t_0}$ using the forward diffusion process in the closed-form~\eq{ddpm_xt}. The edited image is generated for a few steps of the DDIM decoding that sequentially applies the learned transform and recovers the fine-grained details.}
    \label{fig:main_scheme}
\end{figure}

\section{Method}
\label{sect:method}
In this section, we design a more efficient image manipulation algorithm that circumvents the expensive multi-step processing in the DiffusionCLIP training and inference stages.
We conjecture that the objective \eq{objective} does not need precise image reconstruction $x_0(\theta)$ to successfully learn the text-driven image transformation. 
We apply this intuition to speed up training and inference in Sections \ref{sect:ingredient_1} and \ref{sect:ingredient_2}, respectively. \fig{main_scheme} summarizes our approach.

% We conjecture that for successful editing one does not need fine-grained perceptual details to successfully learn the image tranformation when reconstruct images from intermediate diffusion steps. Informally, if an image is used only as an input to a CLIP loss, that image need not to be visually appealing.
% i) expensive adaptation procedure that requires backpropagation through the reverse sampling and% ii) costly sequential inference using the DDIM forward and reverse processes.
% We address each of these problems individually relying on our assumption that fine-grained perceptual details are not crucial for effective domain adaptation. 

\subsection{Single step training}
\label{sect:ingredient_1}

% First, we propose to alleviate the training procedure of DiffusionCLIP described in \sect{diffclip}.
%In DiffusionCLIP, most of the training time is spent on running multi-step sampling procedure and backpropagating through it. 
% Instead, we train the model $\hat{\theta}$ on a single time step $t_0$ without unrolling the DDIM reverse process. 
As we describe in \sect{diffclip}, each DiffusionCLIP training step starts with $x_{t_0}$, applies the DDIM generating process for $\tau_{dec}$ steps to obtain $x_0(\theta)$, then uses it to estimate $\mathcal{L}(\theta)$ and backpropagate through the entire computational graph to update parameters $\theta$.

% To avoid the costly multi-step procedure, we approximate $x_0(\hat{\theta}) \approx x^{t_0}_0(\hat{\theta})$. Instead of sampling an image, the DDIM estimates the expected $x_0$ using eq. \eq{ddim_x0}.

\begin{figure}
    \centering
\includegraphics[width=\columnwidth]{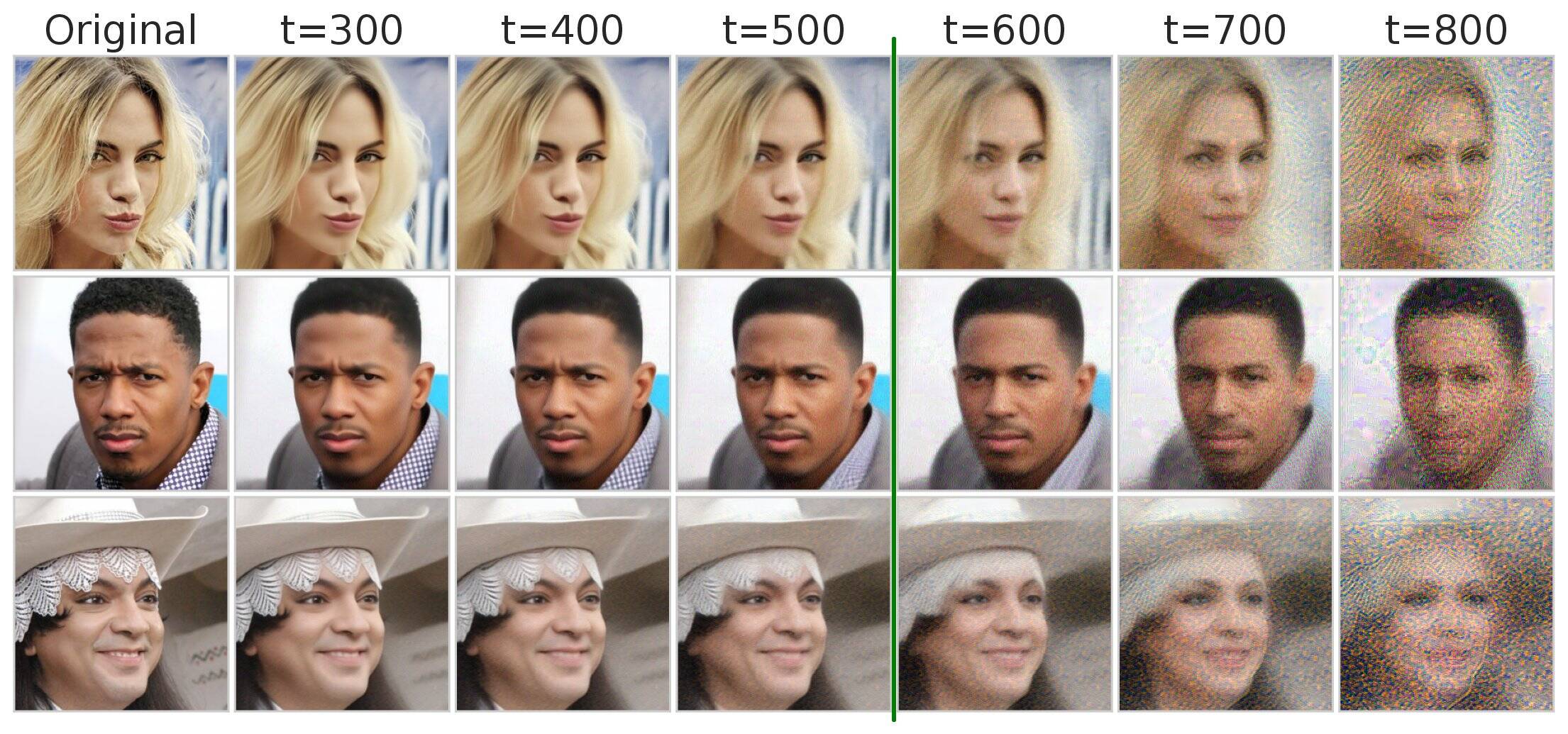}
    \vspace{-6mm}
    \caption{Visualisation of $x_0(t_0; \theta)$ predictions for different $t_0$. The estimates for $t_0\le500$ preserve main semantic attributes of the original images.}
    \label{fig:x0_examples}
\end{figure}

To avoid the costly multi-step procedure, instead of generating an edited image $x_0(\theta)$ for $\tau_{dec}$ steps, we estimate the expected $x_0$ using eq.~\eq{ddim_x0} and denote it as $x_0(t_0; \theta)$. 
The $x_0(t_0; \theta)$ estimate lacks some fine-grained details. 
This is illustrated in \fig{x0_examples} that presents $x(t_0; \theta)$ predictions for different $t_0$ steps\footnote{The predictions are obtained with a diffusion model pretrained on the CelebA-HQ dataset from \url{https://github.com/ermongroup/SDEdit}.}. 
However, we hypothesize that our training procedure only needs $x_0(t_0; \theta)$ to preserve the semantic attributes that are not supposed to be altered during the particular image manipulation. %, e.g., for the ``Makeup'' attribute, we only need to recover relatively low frequency components of a face such as eyes/lips/cheeks, etc.
% While this approach is faster, 

Since $x_0(t_0; \theta)$ is a \emph{single-step} estimate, the method performs only a single forward and backward pass of the diffusion model per training iteration.
This results in both faster training and significantly less memory usage since there is no need to store multi-step activations in GPU memory.
We report the exact runtimes and memory usage in \sect{runtimes}.

% In our experiments, we demonstrate that the omitted high frequency details get successfully recovered by the DDIM decoding at the inference stage. 

%\nikita{but at the inference we do not use single step, so i cannot catch the connection between "high frequency details get successfully recovered by the DDIM sampler at the inference stage" and single step training.}

\textbf{Emergent effect of the training procedure.}
Interestingly, aside from the intended purpose of faster training, the proposed algorithm improves the visual quality of the resulting $x_0(t_0; \hat{\theta})$ predictions.
This effect is not obvious since the objective $\mathcal{L}$ does have any terms that explicitly optimize the image quality.
This phenomenon allows us to reduce the number of DDIM decoding steps $\tau_{dec}$ and further speed up the inference.
In \sect{x0_analysis}, we explore this emergent property in more detail. %\nikita{i think we have to mention that for diffclip this effect does not work cause they consider "exact" reconstructions}

\subsection{Efficient forward processing}
\label{sect:ingredient_2}
In addition, DiffusionCLIP needs the costly DDIM encoding process to precisely invert the sampling procedure up to fine-grained details.
We argue that this exact inversion is unnecessary because semantic image manipulations are performed at the middle steps of the diffusion process where most semantic attributes have already been formed. 
Also, the edit typically requires noticeable image manipulations that override the details generated during the final steps.

Therefore, we suggest replacing the DDIM encoding \eq{ddim} with the DDPM forward process and sample $x_{t_0}{\sim}q(x_{t_0}|x_0)$ in the closed-form using eq.\eq{ddpm_xt}. 
Notably, we use the DDPM forward process only for encoding, retaining DDIM for the decoding phase. 

\begin{figure}
    \centering
    \includegraphics[width=\columnwidth]{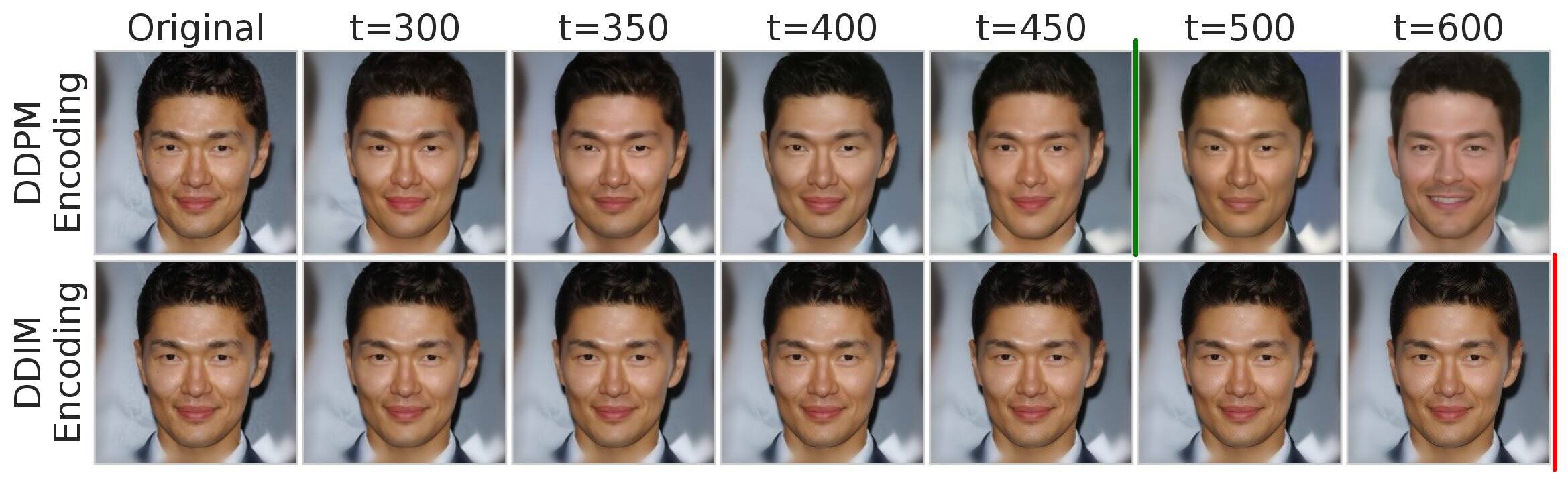}\\
    \includegraphics[width=\columnwidth]{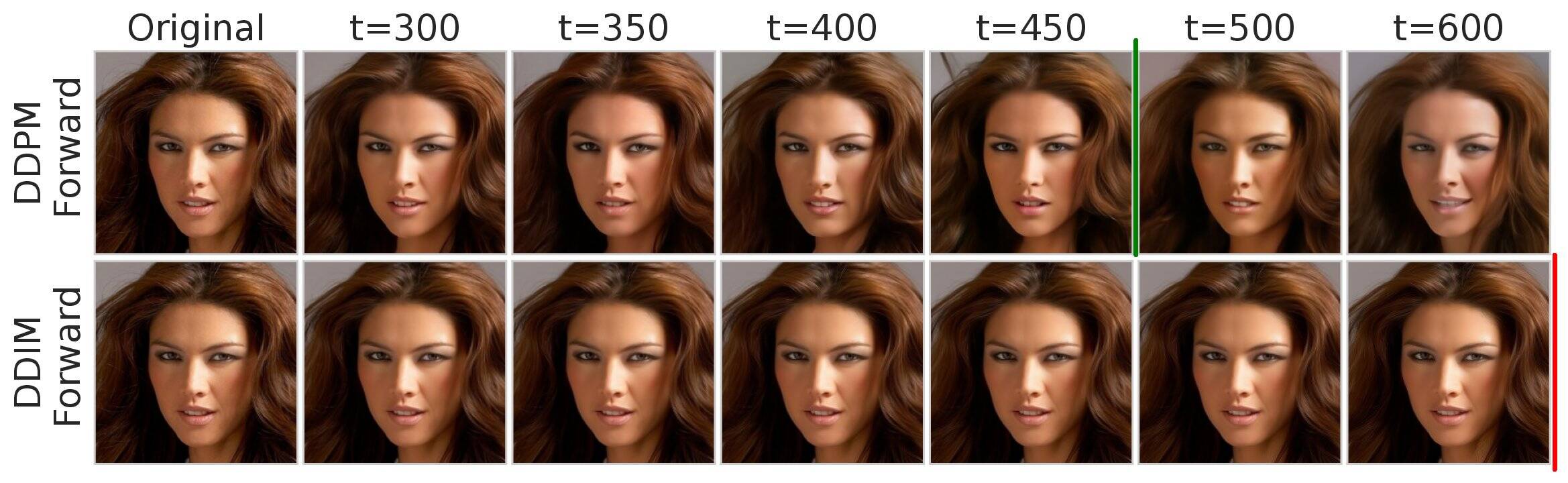}
    \vspace{-6mm}
    \caption{Image reconstruction with stochastic (DDPM) and deterministic (DDIM) encoding methods for different diffusion steps $t_0$. The reconstructions obtained from DDPM encodings for $t_0\le450$ are still consistent with the original images and suit well for most image manipulations.}
    \label{fig:forward_rec}
\end{figure}

In \fig{forward_rec}, we visualize the image reconstructions $x_0(\theta)$ obtained with DDIM and DDPM encoding methods for different $t_0\in[300, 600]$ values. 
We observe that $x_0(\theta)$ reconstructed from $x_{t_0}{\sim}q(x_{t_0}|x_0)$ are still consistent with $x_0$ for $t_0{\le}450$. 
In Appendix~\ref{appendix:mos_encoding}, we provide the mean opinion score that quantitatively supports this observation.  

To sum up, we combine the two modifications above and present the overall training algorithm in \app{training}. 

During the inference, the proposed method encodes an image $x_0$ into the latent variable $x_{t_0}{\sim}q(x_{t_0}|x_0)$ and uses the DDIM decoding for $\tau_{dec}$ steps to gradually apply the learned manipulation and recover high-frequency details. 

\section{Experiments}
In this section, we first investigate the emergent effect caused by our training procedure. Then, we evaluate the efficiency of our approach in comparison with DiffusionCLIP and GAN-based alternatives.
Finally, we demonstrate the qualitative and quantitative results of our approach for the following settings:
\begin{itemize}[leftmargin=12px]
    \item \textbf{Prelearned image manipulations} --- the model is first adapted to the text description on $50$ images. Then, the learned transform is applied to the hold-out images. 
    \item \textbf{Single-image editing} --- the model is adapted to the user-specified text description and image on the fly.
\end{itemize}

% In this section, we evaluate the efficiency of our approach in comparison to DiffusionCLIP and GAN-based alternatives.
% Then, we demonstrate qualitative and quantitative results of our approach for text-driven image translation and single image editing. 
% Finally, we investigate the emergent effect caused by our training procedure and provide the ablation study that highlights the importance of each modification.

\subsection{Analysis}
\label{sect:x0_analysis}

%\textbf{Applicability of the estimation for domain adaptation}. We started our exploration by considering the following basic question: why can the estimation of $x_{0}$ be applied for domain adaptation of diffusion models? To this end, we examine the behaviour of the estimation as a function of $t_{0}$ (noise extent). 

 %Qualitative and quantitative results are presented in Figure \ref{fig:recon_graph}.
%\begin{figure}[h!]
%    \centering
%    \includegraphics[width=\columnwidth]%{images/recon_graph.png}
 %   \caption{\textbf{Results of the reconstruction}. Top: MAE loss between the estimations and the real samples for different $t_{0}$. Bottom: visual representation of the estimation for different $t_{0}$.}
 %   \label{fig:recon_graph}
%\end{figure}
%As it can be seen, the latter demonstrates dual-mode behaviour. That is, there is the change mode point at $t_{0} \approx 450$, after which the reconstruction loss begin to grow significantly. For the high values of $t_{0}$ the estimation shows unsatisfactory quality and thus cannot be applied for domain adaptation. However, usually, semantic editing of images is performed near the change mode point, i.e., in the range $t_{0} \in (300, 600)$ \cite{kim2022diffusionclip, meng2021sdedit}. As it can be seen at the bottom of Figure \ref{fig:recon_graph}, in this range the estimation shows acceptable quality, losing only fine-grained details while saving the most important ones. 

\textbf{Quality improvement effect.}
As we discuss in \sect{ingredient_1}, the proposed training procedure demonstrates an interesting phenomenon: optimizing $\mathcal{L}_{dir}$ improves the perceptual quality of the $x_0(t_0; \hat{\theta})$ estimate after the model adaptation. 
Thanks to this effect, our method can produce samples of the same visual quality for $\tau_{dec}{=}6$ steps instead of $40$ considered in the best DiffusionCLIP configuration.

% the number of the DDIM reverse steps can be reduced. 
% We found that $6$ steps are sufficient for semantic editing, while the best DiffusionCLIP configuration uses $40$.

% In \fig{x0_estim}, we observe that the estimation loses fine-grained details. 
% However, during the optimization they tend to return and the estimation quality becomes closer to the source image, $x_{0}$.

%In \fig{x0_examples}, we visualize a few examples of $\tilde{x}_0(\hat{\theta})$ and $\tilde{x}_0(\theta)$ on the CelebA dataset. 
%We observe that $\tilde{x}_0(\hat{\theta})$ contains more high frequency details compared to $\tilde{x}_0(\theta)$. 

\begin{figure}
% \centering
\hspace{-2mm}
\includegraphics[width=0.49\textwidth]{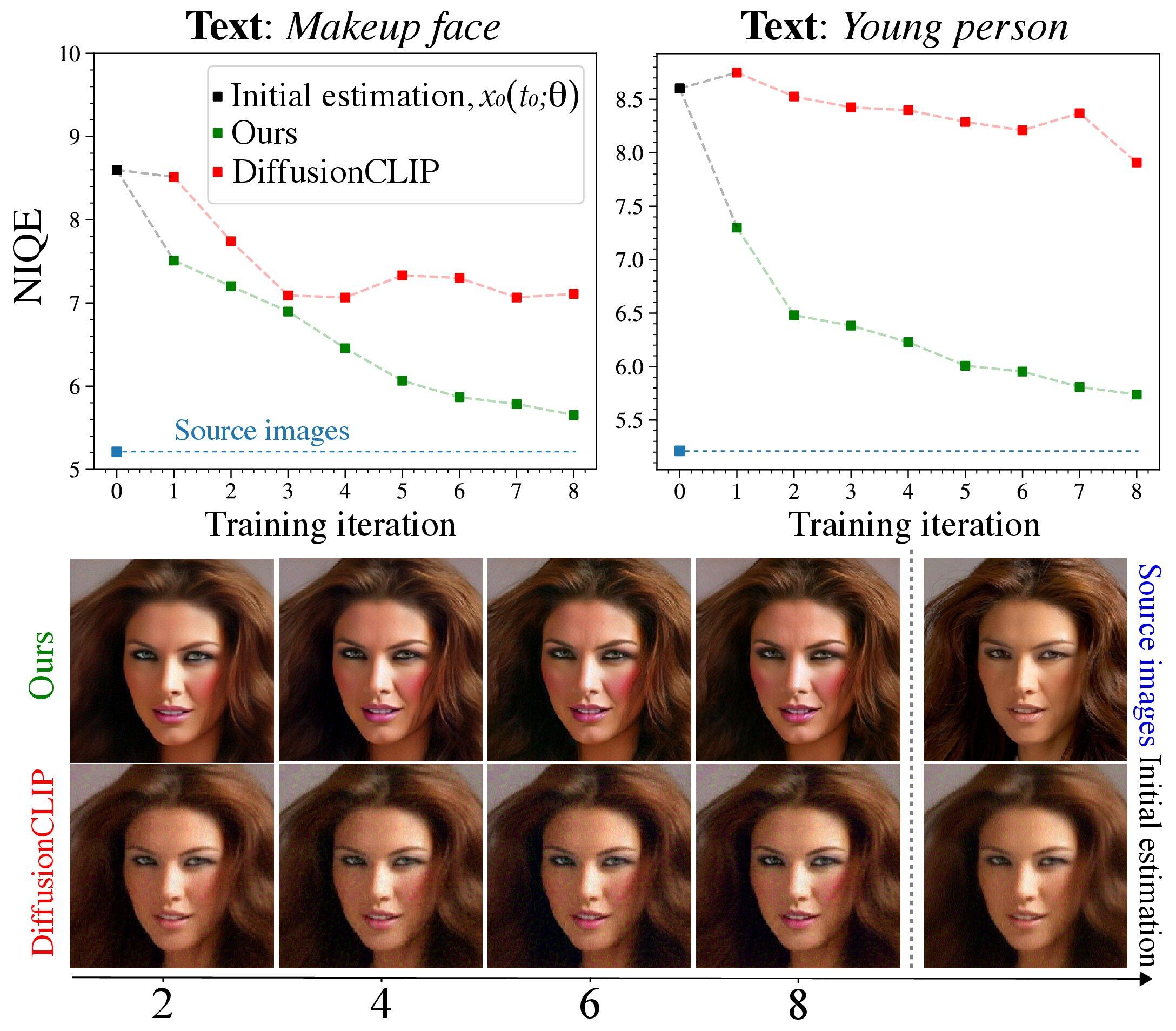}
\vspace{-6mm}
\caption{Image quality (NIQE) of $x_0(t_0; \hat{\theta})$ estimate w.r.t. training iterations. For both methods, the estimates get better but improvements are way more pronounced for the proposed method.}
\label{fig:niqe}
\end{figure}

In \fig{niqe}, we measure the dynamics of $x_0(t_0, \hat{\theta})$ quality w.r.t. different fine-tune iterations using NIQE~\cite{mittal2013niqe} --- the established metric for the no-reference image quality assessment. 
Opposed to DiffusionCLIP, we observe that the quality of the $x_0$ estimates indeed significantly increases over training for our method. 

\textbf{What causes improvement of $x_0(t_0; \hat{\theta})$?}
The directional CLIP loss \eq{clip_loss} aims to move the image embedding of the $x_0(t_0; \hat{\theta})$ estimate to make a vector $E_{I}(x_0(t_0; \hat{\theta})){-}E_{I}(x_0)$ co-directional to the text direction $E_{T}(y_{ref}){-}E_{T}(y_{tar})$ in the CLIP embedding space. 

We hypothesize that the $x_0$ transformations along the text direction weakly correlate to the perceptual quality unless it is assumed by the manipulation. 
If so, the quality of the $x_0(t_0; \hat{\theta})$ is essentially determined by the source image $x_0$. 

\begin{figure}
    \centering
    \includegraphics[width=5.5cm]{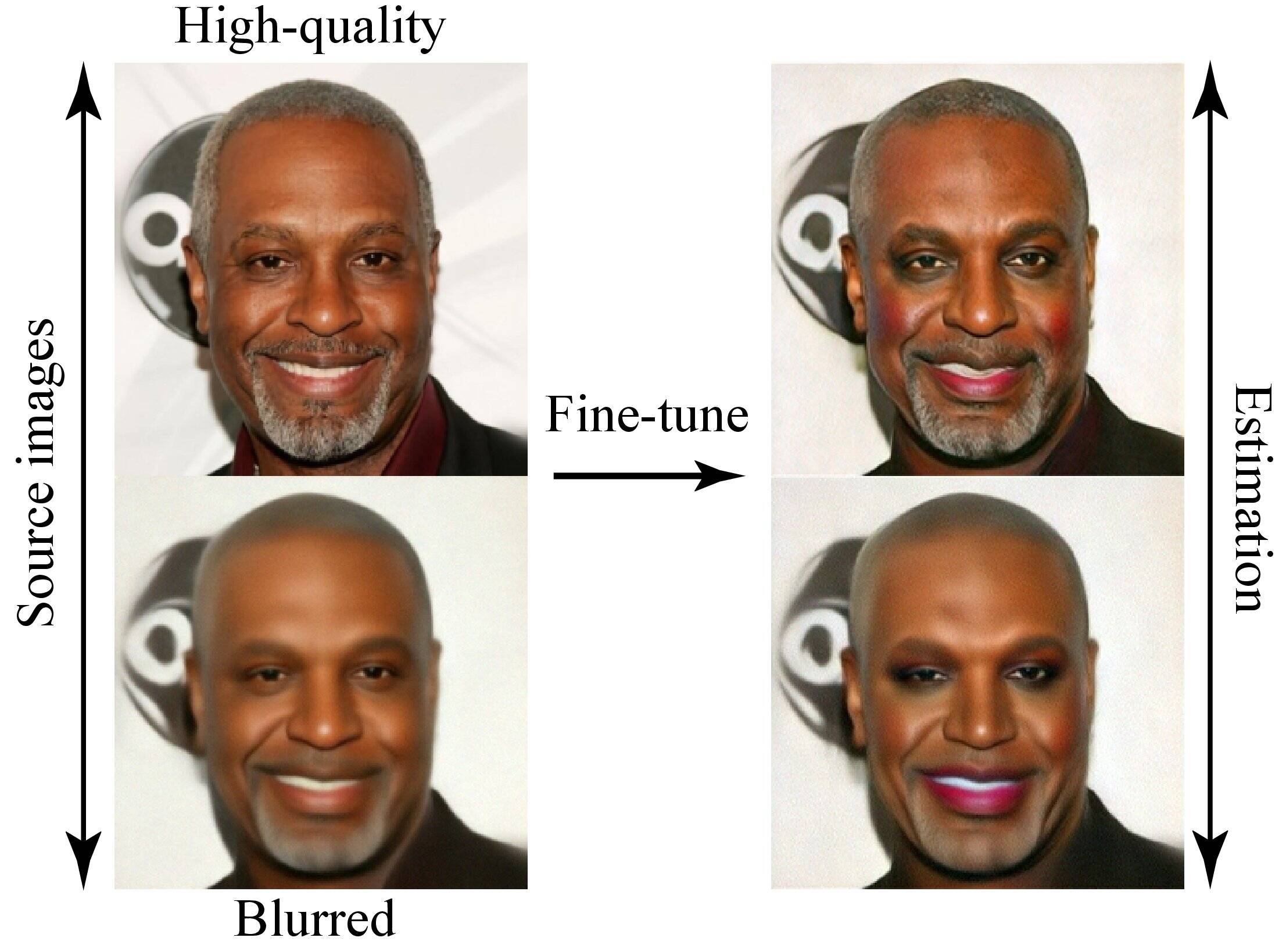}
    \vspace{-2mm}
    \caption{Visualization of $x_0(t_0; \hat{\theta})$ estimates after fine-tuning for $x_0$ of different perceptual quality. The estimates tend to inherit the perceptual details of the source images.}
    \label{fig:quality_eff}
\end{figure}

To confirm our assumption, we first consider some text direction, e.g, ``Face'' to ``Makeup Face'' and optimize $\mathcal{L}_{dir}$ for $x_0$ and its manually blurred version $x^{blur}_0$, independently.
In \fig{quality_eff}, we visualize $x_0(t_0; \hat{\theta})$ estimates after the model adaptation.
We observe that the estimates tend to preserve the sharpness of the source images.
%More visualizations are available in \app{quality_imp}.
 
\begin{figure}
\centering
\includegraphics[width=0.4\textwidth]{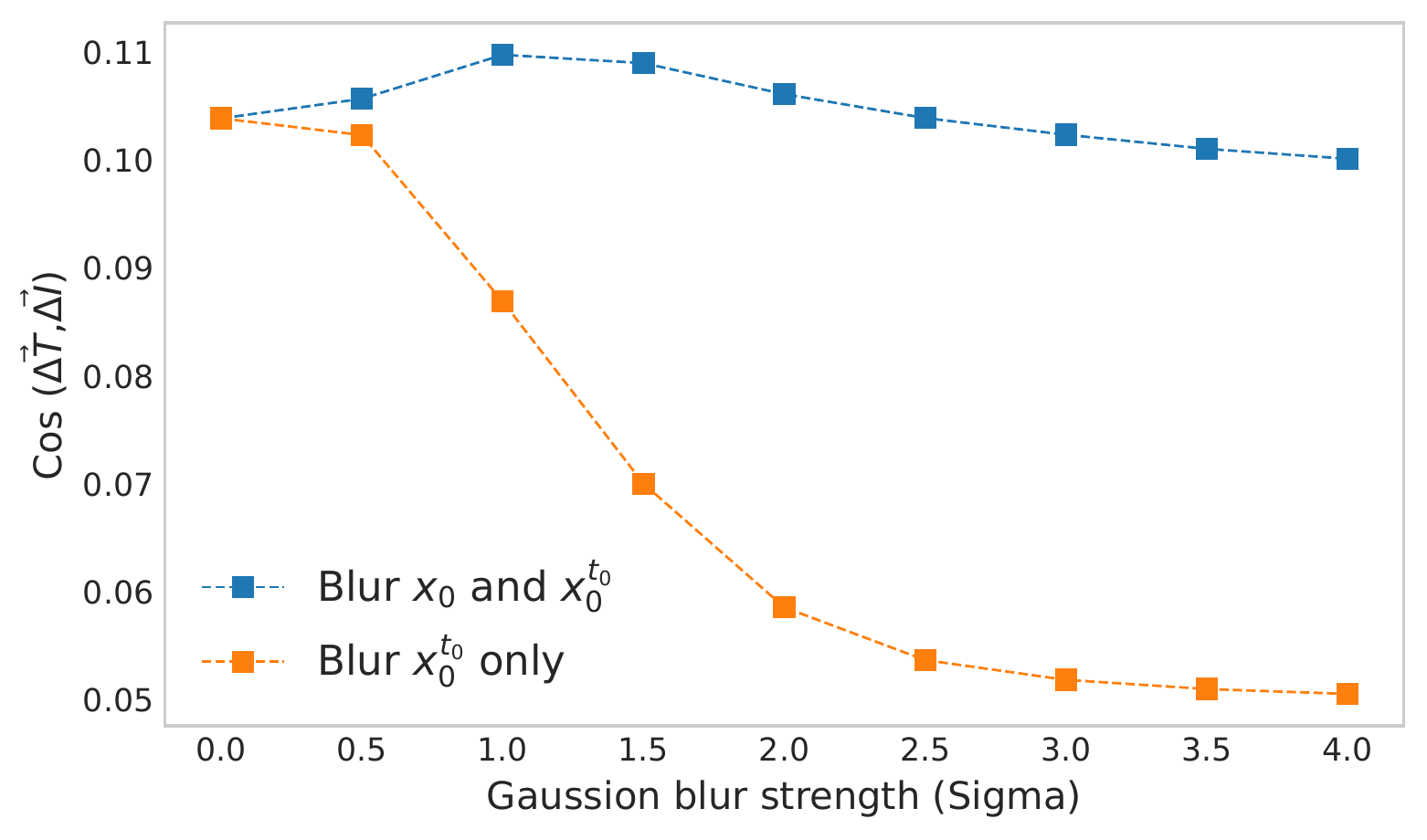}
\vspace{-2mm}
\caption{Cosine between $\Delta T$ and $\Delta I$ directions in the CLIP space w.r.t. Gaussian blur strength applied to the image $x_0$ and its manipulated estimate $x_0(t_0; \hat{\theta})$. The angle is not affected much if both $x_0$ and $x_0(t_0; \hat{\theta})$ are corrupted equally. This indicates that the semantic direction induced by $\Delta T$ has weak correlation with the image quality direction.}
\label{fig:clip_loss_blur}
\end{figure}

Secondly, we explore the behavior of the semantic direction when the perceptual quality of the target and source images changes. 
In more detail, we consider source images $x_0$ and their $x_0(t_0; \hat{\theta})$ estimates after the model adaptation. 
In the first setting, we fix $x_0$ and apply Gaussian blur of different degree only to $x_0(t_0; \hat{\theta})$: $\Delta I(x_0, x^{blur}_0(t_0; \hat{\theta}))$.
In the second setting, we apply Gaussian blur to both $x_0$ and $x_0(t_0; \hat{\theta})$: $\Delta I(x^{blur}_0, x^{blur}_0(t_0; \hat{\theta}))$.
We consider Gaussian blur with kernel size $7$ and vary sigma from $0$ to $4$.
Then, we measure cosine between the image direction $\Delta I$ and the text direction $\Delta T$ for various $y_{tar}$.
The results are averaged over $6$ transforms and presented in \fig{clip_loss_blur}.
We observe that the image direction ${\Delta}I$ does not noticeably change if both images are equally corrupted. 

%Finally, we address whether $\mathcal{L}_{direction}$ favors higher quality images from the target domain.
%Since there are no reference images from the target domain, we transform a source image $x_0$ using $\epsilon_{\hat{\theta}}$ and use the generated $x_0(\hat{\theta})$ instead.
%Then, we corrupt $x_0(\hat{\theta})$ by applying Gaussian blur with kernel size $5$ and sigma $1$ and compute $\mathcal{L}_{direction}$ for $x_0(\hat{\theta})$ and $x^{blurred}_0(\hat{\theta})$.
%In \tab{}, we compare average $\mathcal{L}_{direction}$ over various target domains $y_{tar}$ on the CelebA, ImageNet and AFHQ datasets.
%Indeed, $\mathcal{L}_{direction}$ is lower for higher quality $x_0(\hat{\theta})$ samples.

\subsection{Training and inference efficiency}
\label{sect:runtimes}
%The main application of our approach is text-driven domain adaptation of unconditional diffusion models. We suppose that any approach to this problem for its successful practice utilization must meet two key conditions: \textit{efficient training and evaluation} and \textit{high quality of transformations}. However, modern approaches cannot satisfies both criteria. On the one hand, the DiffusionCLIP that suffers from inefficient training and evaluation. On the other hand, the StyleGAN-NADA that demonstrates worse results than the diffusion based approach. The key idea of this work is to try to take the best of both worlds. 

% In this section, we investigate runtimes and memory consumption of the proposed method and compare it to the DiffusionCLIP approach. 
% At the end, we also compare the overall method against the state-of-the-art GAN-based methods for domain adaptation (StyleGAN-NADA~\cite{gal2021stylegannada}) and single image editing (StyleCLIP~\cite{Patashnik_2021_ICCV}.

\textbf{Setting.}
The proposed method and DiffusionCLIP exploit the same pretrained models.
As a GAN-based baseline, we consider StyleGAN-NADA~\cite{gal2021stylegannada} that is built upon the StyleGAN2~\cite{Karras2019stylegan2} generator.
In this evaluation, all models are pretrained on CelebA-HQ and operate on $256{\times}256$ images. 
% The particular checkpoints are publicly available\footnote{https://github.com/ermongroup/SDEdit}.
% In this experiment, we also consider the official checkpoints on CelebA-HQ for the $256{\times}256$ resolution\footnote{TODO}. \dmitry{confirm that it is true for both e4e and stylegan}

All measurements are performed in an isolated environment on a single NVIDIA A100 GPU and averaged over $10$ independent runs.
For both baselines, we consider the official implementations and the best settings described in the corresponding papers. 
We provide the exact hardware and software configuration in \app{tech_setup}.
% We begin our investigation by examining the effectiveness of the training and evaluation phases of our approach. In a nutshell, we conducted the following experiments: 1) a comparison of the training efficiency between the proposed approach and DiffusionCLIP in terms of time and memory costs; 2) a comparison of the inference time between our method and the diffusion/GAN based approaches.

\begin{figure}
    \centering
    \includegraphics[width=\columnwidth]{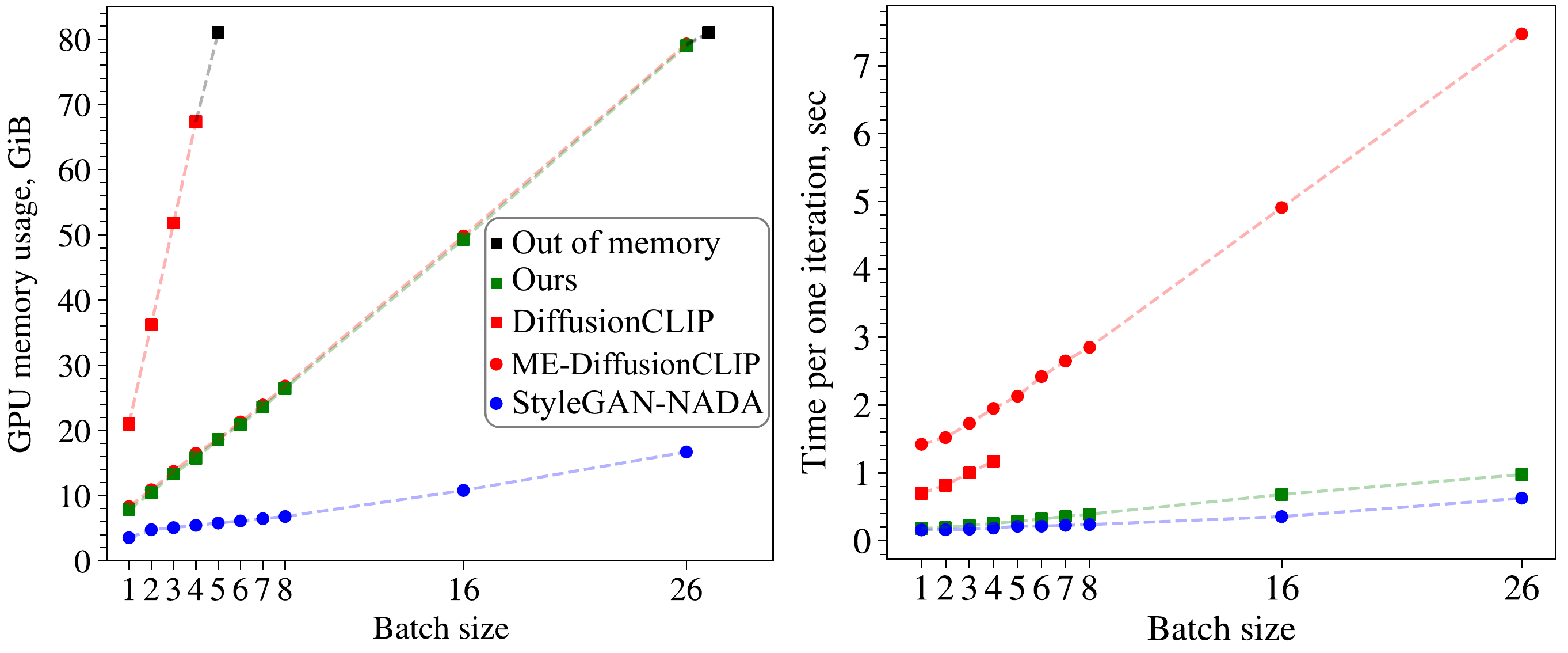}
    \vspace{-4mm}
    \caption{Comparison of our approach, DiffusionCLIP and StyleGAN-NADA in terms GPU memory usage (GiB) (Left) and run-times (sec) (Right) for a single training iteration.}
    \label{fig:timegpu1}
\end{figure}

\begin{table}
\centering
\resizebox{6.5cm}{!}{
 \begin{tabular}{|c| c c |} 
 \hline
 Method & Encoding, sec  & Decoding, sec  \\ [0.5ex] 
 \hline
  StyleGAN-NADA & $0.29$ & $0.07$ \\   
 DiffusionCLIP & $1.53$ & $0.22$ \\ 
 Ours & $0.00$ & $0.22$ \\[1ex] 
 \hline
 \end{tabular}%
 }
 \vspace{-2mm}
 \caption{The time spent to encode and decode a single image of $256{\times}256$ resolution at the inference stage.}
\label{tab:time_mes}
\vspace{-3mm}
\end{table}

\textbf{Training efficiency}. 
We start with the evaluation of the training performance of each method.
The training set contains $50$ images.

The first phase is to precompute the latent variables $x_{t_0}$ for all training images $x_0$. 
DiffisionCLIP exploits the DDIM encoding for $\tau_{enc}{=}40$ steps and spends ${\sim}67$ seconds to encode $50$ images. 
On the other hand, our method gets $x_{t_0}$ in the closed form.
StyleGAN-NADA is trained on self-generated samples and hence skips this phase.

Then, in \fig{timegpu1}, we compare the GPU memory consumption (Left) and runtime (Right) of a single training iteration w.r.t. different batch sizes.
As an additional baseline, we consider the memory efficient (ME) DiffusionCLIP procedure described in \sect{diffclip}. 
During the training, DiffusionCLIP performs $\tau_{gen}{=}6$ decoding steps.

We observe that our approach is significantly faster than both DiffusionCLIP versions.
For example, our method demonstrates $0.18$s against $0.70$s and $1.42$s for DiffusionCLIP and ME-DiffusionCLIP, respectively. 
Notice that the gap increases with a batch size.

In terms of GPU memory usage, our procedure consumes $7$GiB for a batch size $1$ that is $14$GiB less compared with the original DiffusionCLIP method.
Moreover, DiffusionCLIP exceeds the GPU memory limit ($>80$GiB) for a batch size $5$ while our method consumes only ${\sim}18$GiB and meets OOM only for a batch size $27$.
ME-DiffusionCLIP consumes similar GPU memory to our method because both perform a model update only for a single diffusion step at a time.
On the contrary, StyleGAN-NADA is still noticeably more memory efficient than diffusion-based alternatives, especially for large batch sizes.

Finally, we measure the overall training performance.  
DiffusionCLIP and our method perform $1{-}10$ training epochs depending on the target transform.
Thus, for a batch size $1$, our procedure spends from $10.7$ seconds to $1.58$ minutes, while the fastest DiffusionCLIP procedure spends from $101$ seconds to $7.2$ minutes.
Note that the efficiency gains increase for larger batch sizes.

StyleGAN-NADA requires $50{-}300$ training iterations and operates on a batch size $2$.
In this case, the overall training takes from $8.1$ seconds to $58.5$ seconds. 

In the result, our training procedure is $4.5{-}10{\times}$ faster and consumes $14$GiB less GPU memory than original DiffusionCLIP and $8{-}18{\times}$ faster than its memory efficient alternative. 
In comparison with StyleGAN-NADA, our adaptation still consumes extra $4$ GiB memory but, notably, is comparable in terms of the training time.

\begin{figure*}
\label{fig:domain_view}
    \centering
    \vspace{-2mm}
    \includegraphics[width=17.6cm]{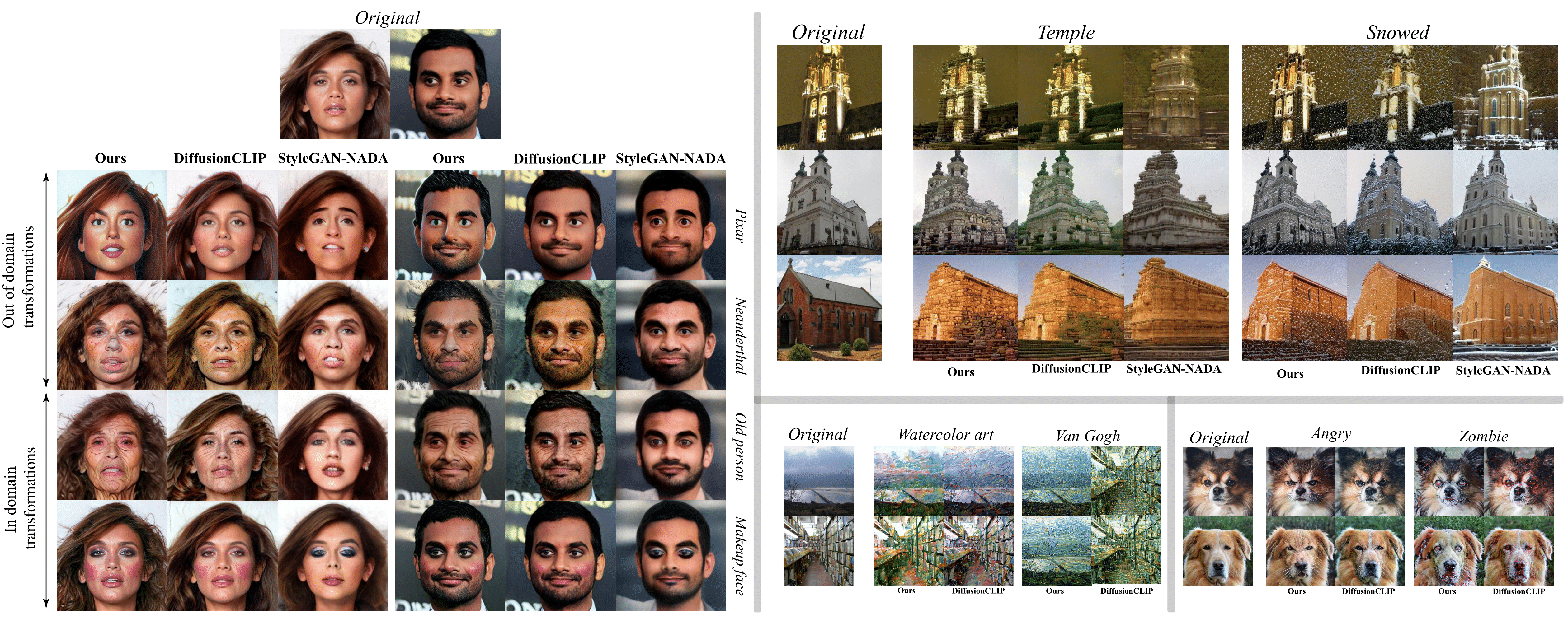}
    \vspace{-8mm}
    \caption{Visual examples of the image manipulations learned with the DiffusionCLIP, StyleGAN-NADA and our approaches for the Celeba-HQ-256, LSUN-church-256, AFHQ-dog-256 and ImageNet-512 datasets.}
    \label{fig:domain_view}
\end{figure*}

\textbf{Inference time}. 
Then, we compare the method performance at the inference stage.
The fastest DiffusionCLIP configuration considers $\tau_{enc}{=}40$ encoding steps and the $\tau_{dec}{=}6$ for decoding.
Our approach encodes an image for free and performs $\tau_{dec}{=}6$ decoding steps.
StyleGAN-NADA utilizes Restyle~\cite{alaluf2021restyle} + e4e~\cite{tov2021designing} as an encoding method. 

The results are presented in \tab{time_mes}.
As one can see, our approach with efficient approximate encoding results in $8{\times}$ faster inference than DiffusionCLIP. 
Moreover, it also demonstrates ${\sim}50\%$ speedup compared to StyleGAN-NADA due to the costly image encoder.

\subsection{Prelearned image manipulations}
\label{sect:main_exp}
For this setting, we evaluate the quality of the image manipulations on the Celeba-HQ-256~\cite{karras2018progressive}, AFHQ-dog-256~\cite{choi2020starganv2}, LSUN-Church-256~\cite{yu15lsun} and ImageNet-512~\cite{DenDon09Imagenet} datasets. 
As baseline methods, we consider DiffusionCLIP, StyleGAN-NADA, Asyrp~\cite{kwon2023diffusion} and HyperDomainNet~\cite{alanov2022hyperdomainnet}.
For each method, we consider the hyperparameter values from the corresponding official implementations and papers.
If some hyperparameters are missing, we carefully tune them by ourselves. 
For DiffusionCLIP, opposed to the fastest setting with $\tau_{dec}{=}6$ in \sect{runtimes}, in this experiment, we set $\tau_{dec}{=}40$ during the inference to derive the best editing results.
Note that our approach still performs $\tau_{dec}{=}6$ decoding steps.  
For comparison, we use $35$ images from the test sets. 
$16$ text descriptions are taken for CelebA-HQ-256 and $6$ for other datasets. 
The list of textual transforms is in \app{domain_transform_list}.
% To collect the data, we first train the baselines using their official implementations and the proposed approach with a configuration similar to the DiffusionCLIP. Then, we applied the fine-tuned models to 35 unseen samples from the test datasets. 
% We just mention that our evaluation process uses six DDIM steps against 40 of DiffusionCLIP because of the quality improvement effect.

As a primary quality measure, we consider a side-by-side human evaluation.
Specifically, we provide a source image, text description and images edited with two different methods and then ask people to answer two questions: 1)\textit{Which of the edited images corresponds better to the text description?} and 2)\textit{Which of the edited images has more artifacts and changes that are not related to the text?}. 
In total, we collect $9500$ votes. 
The voting results are presented in \tab{our_diffclip}. 
\fig{domain_view} provides the visualizations for various transformations.
More visual examples are in Figures \ref{fig:comparison_adapt}, 
\ref{fig:comparison_semantic},
\ref{fig:comparison_dogs}, \ref{fig:comparison_imagenet}, \ref{fig:comparison_church}, \ref{fig:comparison_gans}. %\app{extra_results}.
The qualitative comparison with HyperDomainNet~\cite{alanov2022hyperdomainnet} is presented in \fig{comparison_hyper}.

Compared with DiffusionCLIP, our approach corresponds better to the text attributes on Celeba-HQ and LSUN-Church.
On other datasets, the votes are distributed equally.
In terms of artifacts, we observe parity between both methods.
This means that our method still produces high-fidelity images and provides as expressive image transforms as DiffusionCLIP. 
Moreover, despite using stochastic encoding, our algorithm does not induce noticeable artifacts in semantic attributes: even for the shallow transformations, e.g., ``Makeup face'', images still preserve important source details.

Against StyleGAN-NADA, Asyrp and HyperDomainNet, the proposed method significantly outperforms all of them according to both criteria.

\begin{figure*}
    \centering
    %\vspace{6mm}
    %\hspace{-2mm}
    \includegraphics[width=1.98\columnwidth]{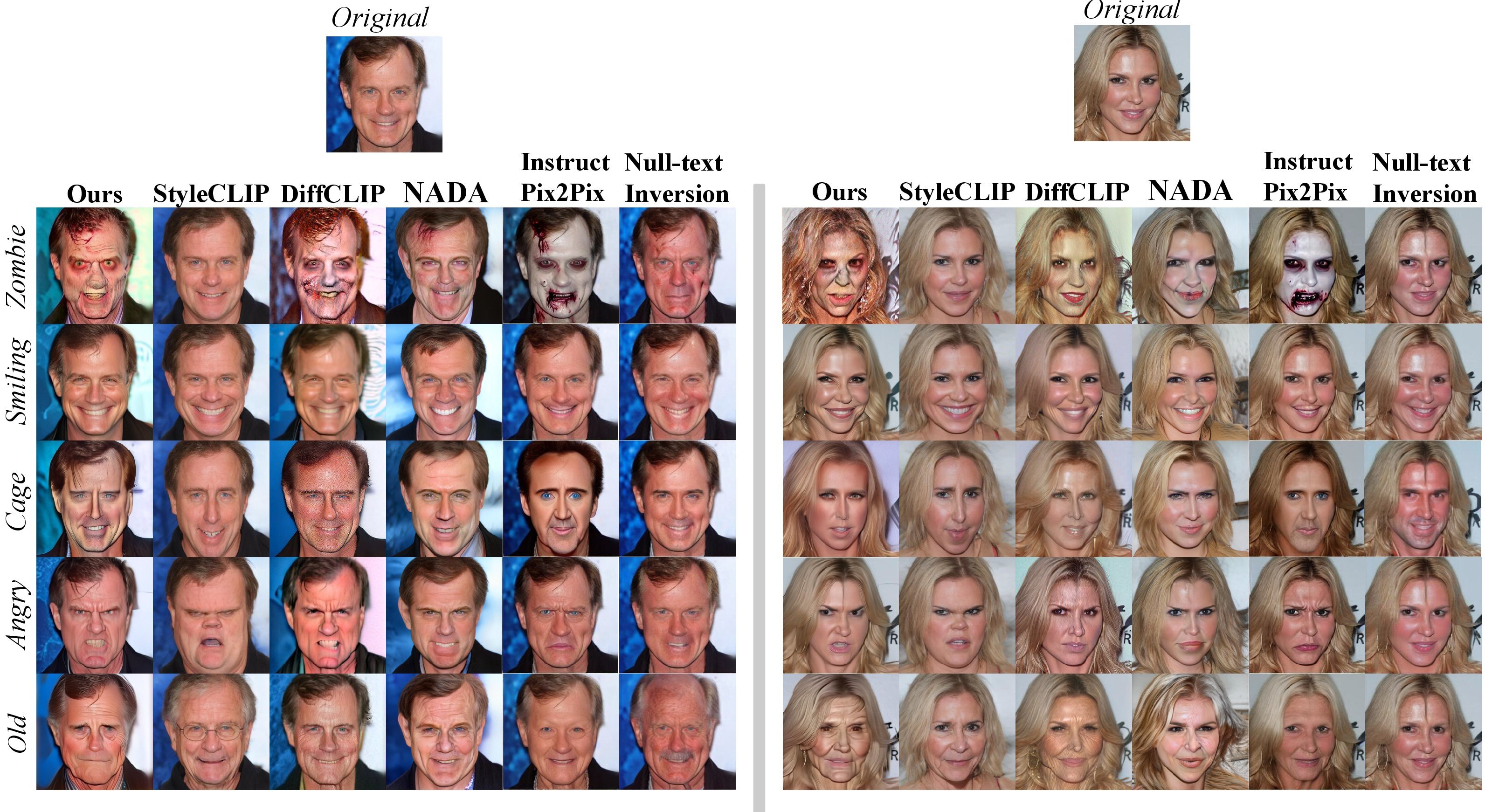}
    \vspace{-4mm}
    \caption{Visual examples produced with different single-image editing approaches. 
    Our method, DiffusionCLIP, StyleCLIP and StyleGAN-NADA represent the methods using unconditional generative models. InstructPix2Pix and Null-text Inversion methods are based on Stable Diffusion.}
    \label{fig:single_compar}
\end{figure*}

\begin{table}
\centering
\resizebox{8.5cm}{!}{%
 \begin{tabular}{|c| c c c|} 
 \hline
 Dataset & Ours, \%  & Both, \%   & DiffusionCLIP, \%  \\ [0.5ex] 
 \hline
 &\multicolumn{3}{c|}{\textit{Q1}. Which one corresponds better to the text?}  \\ [1ex] 
  Celeba-HQ & $\mathbf{54.21} \pm 1.82$ & $5.96\pm 1.23$ & $39.82\pm 2.68$ \\   
 AFHQ-Dog & $\mathbf{44.11} \pm 3.25$ & $16.74 \pm 4.65$ & ${38.85} \pm 2.12$ \\
 ImageNet & $35.42\pm 4.11$  & $31.23\pm 1.96$ & $33.23\pm 3.37$  \\ 
  LSUN-Church & ${40.67} \pm 3.85$ & $18.95 \pm 1.39$ & ${40.38} \pm 3.22$ \\  [1ex] 
  &\multicolumn{3}{c|}{\textit{Q2}. Which one has more text-irrelevant changes?} \\ [1ex] 
 Celeba-HQ & ${40.53} \pm 1.89$ & $20.60\pm 2.05$ & ${38.85}\pm 1.95$ \\  
 AFHQ-Dog &  $40.28\pm 3.53$ & ${21.43} \pm 4.52$ & ${37.70} \pm 3.07$ \\
 ImageNet & $39.90 \pm 4.19$  & $21.90 \pm 4.71$ & $38.19 \pm 4.62$ \\ 
 LSUN-Church & ${35.23} \pm 1.78$ & $28.28 \pm 1.98$ & ${36.47} \pm 1.98$ \\[1ex]
 \hline\hline
 Dataset & Ours, \%  & Both, \%   & StyleGAN-NADA, \%  \\ [0.5ex] 
 \hline
&\multicolumn{3}{c|}{\textit{Q1}. Which one corresponds better to the text?} \\ [1ex] 
Celeba-HQ & $\mathbf{57.26} \pm 1.65$ & $6.65\pm 0.20$ & $35.95\pm 1.61$ \\   
 LSUN-Church & $\mathbf{69.04} \pm 3.75$ & $9.05 \pm 1.85$ & $21.23 \pm 2.67$ \\  [1ex]
&\multicolumn{3}{c|}{\textit{Q2}. Which one has more text-irrelevant changes?} \\ [1ex] 
Celeba-HQ & ${29.06} \pm 2.06$ & $21.79\pm 2.17$ & $\mathbf{48.89}\pm 0.72$ \\   LSUN-Church & ${18.76} \pm 3.28$ & $19.23 \pm 1.82$ & $\mathbf{62.01} \pm 3.71$ \\[1ex]
 \hline \hline
  Dataset & Ours, \%  & Both, \%   & Asyrp, \%  \\ [0.5ex] 
 \hline
&\multicolumn{3}{c|}{\textit{Q1}. Which one corresponds better to the text?} \\ [1ex] 
Celeba-HQ & $\mathbf{81.50} \pm 6.6$ & $5.53\pm 1.92$ & $13.05\pm 5.83$ \\    [1ex]
&\multicolumn{3}{c|}{\textit{Q2}. Which one has more text-irrelevant changes?} \\ [1ex] 
 Celeba-HQ & ${21.50} \pm 5.36$ & $22.06\pm 6.00$ & $\mathbf{57.01}\pm 8.77$ \\    [1ex]
  \hline \hline
  Dataset & Ours, \%  & Both, \%   & HyperDomainNet, \%  \\ [0.5ex] 
 \hline
&\multicolumn{3}{c|}{\textit{Q1}. Which one corresponds better to the text?} \\ [1ex] 
Celeba-HQ & $\mathbf{56.72} \pm 3.13$ & $2.53\pm 2.02$ & $40.83\pm 1.34$ \\    [1ex]
&\multicolumn{3}{c|}{\textit{Q2}. Which one has more text-irrelevant changes?} \\ [1ex] 
 Celeba-HQ & ${36.75} \pm 1.56$ & $22.50\pm 4.33$ & $\mathbf{40.81}\pm 2.24$ \\    [1ex]
\hline
 \end{tabular}}
 \vspace{-2mm}
 \caption{Side-by-side human evaluation of various image manipulations learned with the proposed, DiffusionCLIP, Asyrp and StyleGAN-NADA approaches. Our method preserves quality of DiffusionCLIP and significantly outperforms other methods according to both criteria.}
 \label{tab:our_diffclip}
\end{table}

\begin{table}
\centering
\centering
\resizebox{8.5cm}{!}{%
 \begin{tabular}{|c| c c c|} 
 \hline
 &Ours , \%& Both, \% & DiffusionCLIP, \%  \\ [1ex] 
  Text correspondence & $\mathbf{60.20} \pm 3.80$ & $5.05\pm 2.30$ & $34.74\pm 5.81$ \\ 
  More irrelevant changes & ${36.56} \pm 4.06$ & $26.46\pm 4.39$ & ${36.76}\pm 5.73$ \\ [1ex] 
 \hline\hline
  &Ours, \% & Both, \% & StyleCLIP, \%  \\ [1ex] 
 Text correspondence & $\mathbf{53.33} \pm 6.40$ & $10.41\pm 3.49$ & $36.25\pm 8.89$ \\ 
 More irrelevant changes & $27.50 \pm 4.25$ & $23.33\pm 6.89$ & $\mathbf{49.17}\pm 3.86$ \\[1ex]
  \hline\hline
  &Ours, \% & Both, \% & StyleGAN-NADA, \%  \\ [1ex] 
 Text correspondence & $\mathbf{71.57} \pm 5.22$ & $7.84\pm 2.70$ & $20.58\pm 3.82$ \\ 
 More irrelevant changes & $35.68 \pm 3.59$ & $19.21\pm 3.73$ & $\mathbf{45.09}\pm 2.40$ \\[1ex]
 %  \hline\hline
 %  &Ours, \% & Both, \% & StyleCLIP-SG3, \%  \\ [1ex] 
 % Text correspondence & $\mathbf{61.32} \pm 8.62$ & $7.34\pm 5.70$ & $31.37\pm 9.32$ \\ 
 % More irrelevant changes & $38.74 \pm 3.21$ & $14.71\pm 10.03$ & $\mathbf{46.72}\pm 5.11$ \\[1ex]
   \hline\hline
  &Ours, \% & Both, \% & Null-text Inversion, \%  \\ [1ex] 
 Text correspondence & $\mathbf{68.32} \pm 6.85$ & $5.81\pm 2.00$ & $25.81\pm 3.82$ \\ 
 More irrelevant changes & $42.31 \pm 6.71$ & $16.84\pm 8.73$ & ${40.82}\pm 8.62$ \\[1ex]
   \hline\hline
  &Ours, \% & Both, \% & InstructPix2Pix, \%  \\ [1ex] 
 Text correspondence & $36.88 \pm 4.50$ & $6.25 \pm 2.38$ & $\mathbf{56.88} \pm 4.30$ \\ 
 More irrelevant changes & $\mathbf{52.63} \pm 4.47$ & $16.32\pm 1.97$ & ${31.05}\pm 2.86$ \\[1ex]
 \hline
 \end{tabular}
 }
 \vspace{-2mm}
 \caption{Side-by-side human evaluation for single-image editing on CelebA-HQ-256. Our method outperforms DiffusionCLIP and Null-text Inversion according to text correspondence and GAN-based methods according to both criteria. InstructPix2Pix provides better editing results for most transforms.}
 \label{tab:single_votes}
\end{table}

\subsection{Single-image editing} 
\label{sect:single-img}
In this experiment, we make use of a significantly more efficient training procedure and consider our approach for text-driven single-image editing.

In this setting, we consider CelebA-HQ-256 and compare our method with DiffusionCLIP, StyleGAN-NADA and StyleCLIP~\cite{Patashnik_2021_ICCV} --- GAN-based approach that allows fast single-image editing.
% Following~\cite{alaluf2022times}, we also consider StyleCLIP based on StyleGAN3 generator~\cite{Karras2021} and denote it as StyleCLIP-SG3.

In addition, popular text-conditional models also offer the tools to edit a single image for the user-specified text description. 
Therefore, we compare our method with recent editing methods~\cite{mokady2022nulltext, brooks2022instructpix2pix} based on Stable Diffusion~\cite{rombach2022high}. 

% Note that, in this setting, the image encoding costs are negligible compared to the model adaptation time.
% Therefore, our method can deal with both deterministic (DDIM) and stochastic (DDPM) encodings. 
% In our experiments, we still use the DDPM encoding for consistency and observe that the deterministic one brings similar results.

First, we measure the overall time to apply the transform to a source image.   
DiffusionCLIP and our approach require ${\sim}20$ training iterations for high quality adaptation.
StyleGAN-NADA starts producing reasonable image transforms for $150$ training iterations.
Overall, our method, DiffusionCLIP, StyleGAN-NADA and StyleCLIP demonstrate $3.8$, $14.9$, $35.8$ and $1.1$ seconds per image, respectively. 
Null-text inversion~\cite{mokady2022nulltext} requires $40{-}120$ seconds to edit a single image. 
On the other hand, InstructPix2Pix~\cite{brooks2022instructpix2pix} performs text-guided editing for a single forward pass of Stable Diffusion which takes ${\sim}9$ seconds.

Then, we evaluate the visual quality of the manipulated images. 
We consider $6$ images and $8$ in-domain and $8$ out-of-domain textual transforms. 
The full list of textual transforms is in \app{sin_transform_list}. 
The results for some of them are presented in \fig{single_compar}. 
More examples for various textual transforms are in Figures~\ref{fig:sin_images}, ~\ref{fig:sin_pix2pix}.

Also, we run the side-by-side comparison collecting $1860$ votes for all evaluations and present the results in \tab{single_votes}.
Note that StyleCLIP is not designed to handle out-of-domain manipulations, e.g., ``Person'' to ``Zombie''.
Therefore, for a fair evaluation, we compare our method to StyleCLIP only on the in-domain transformations.

We observe that our training procedure demonstrates higher text correspondence compared to DiffusionCLIP under the same level of artifacts and significantly outperforms all GAN-based methods according to both criteria.

For Null-text Inversion, we also select $6$ out of $16$ transforms for which it is able to produce reasonable manipulation results. 
For these transforms, our method still provides more expressive results under the same level of text-irrelevant changes.
%In \app{single_res}, we provide a few examples of the failed transforms (e.g., ``Botero'' and ``Neanderthal'') that we do not take into account during the human evaluation.
% According to both qualitative and quantitative results, we conclude that our approach outperforms Null-text Inversion and, in addition, is an order of magnitude faster.

Our strongest competitor, InstructPix2Pix demonstrates impressive results for most transforms in our evaluation. 
However, we notice that there are some transforms where InstructPix2Pix produces either too weak or too unrealistic results, e.g., ``Smiling person'', ``Surprised person'' or ``Makeup face''.\footnote{We followed the tips provided by the authors but still could not get plausible results.} Therefore, if one wants to add new manipulations, one needs to reproduce the entire pipeline of InstructPix2Pix training that includes incredibly expensive data collection and adaptation procedures. 
In contrast, our method uses the pretrained diffusion model out-of-the-box and needs only a single image to learn the desired transform. 
Notably, our approach provides reasonable image manipulations for all considered transforms. More visual examples in comparison with InstructPix2Pix are in \fig{sin_pix2pix}.

\section{Conclusion}
This work addresses the inefficiency of existing image manipulation approaches built upon unconditional diffusion models. 
We observe that image manipulations can be efficiently learned on inaccurate source image predictions without noticeable loss in perceptual and editing quality. 
In the result, we obtain a highly performant and memory-efficient procedure that can be further considered for real-world applications looking for expressive and efficient image manipulation methods.

{\small
\bibliographystyle{ieee_fullname}
\bibliography{egbib}
}

\clearpage
\newpage

{
 \Large \bf Appendix\\[0.2em]
 \vspace{-2mm}
}

\appendix

\section{Training algorithm}
\label{appendix:training}

(\textcolor{red}{Red}) represents the snippets of the DiffusionCLIP procedure that are replaced with the proposed modifications in \sect{ingredient_1} and \sect{ingredient_2} (\textcolor{dartmouthgreen}{Green}).

\begin{algorithm}
\caption{Training procedure. \textcolor{red}{DiffusionCLIP} vs \textcolor{dartmouthgreen}{Ours}.}
\scriptsize
\DontPrintSemicolon
\SetAlgoLined
    \SetKwInOut{Input}{Input}\SetKwInOut{Output}{Output}
    \Input{$\epsilon_{\theta}, n_{iter}, x_0, y_{ref}, y_{tar}, \tau_{enc}, \tau_{dec}$}
    \Output{$\epsilon_{\hat{\theta}}$}
    \tcc{Encoding process}
    \textcolor{red}{\For{$\tau = 0, ..., \tau_{enc}$}{
       $x_0(\tau; \theta)=x_{\tau}{-}\sqrt{1{-}\alpha_{\tau}}{\cdot}\epsilon_{\theta}(x_{\tau})/\sqrt{\alpha_{\tau}}$\\
        $x_{\tau+1}(\theta)=\sqrt{\alpha_{\tau+1}}{\cdot} x^{\tau}_0(\theta){+}\sqrt{1{-}\alpha_{\tau+1}}\epsilon_{\hat{\theta}}(x_{\tau})$
    }}
    \textcolor{dartmouthgreen}{
        $x_{t_0}{=}\sqrt{\bar{\alpha}_t}{\cdot}x_0 + \sqrt{1-\bar{\alpha}_t}{\cdot}I$ \\   
    }
    $\hat{\theta} = \theta$ \\
    \For{$i = 1, ..., n_{iter}$}{
        %\tcc{DDIM algorithm to obtain $I_{1}$ (denoised images)}
        \tcc{Decoding process and model update}
        \textcolor{red}{\For{$\tau = \tau_{dec}, ..., 1$}{
            $x_0(\tau; \hat{\theta})=x_{\tau}{-}\sqrt{1{-}\alpha_{\tau}}{\cdot}\epsilon_{\hat{\theta}}(x_{\tau})/\sqrt{\alpha_{\tau}}$\\
            $x_{\tau-1}(\hat{\theta})=\sqrt{\alpha_{\tau-1}}{\cdot} x^{\tau}_0(\hat{\theta}){+}\sqrt{1{-}\alpha_{\tau-1}}\epsilon_{\hat{\theta}}(x_{\tau})$
        }
        $\hat{\theta} = \hat{\theta} + \nabla_{\hat{\theta}} \mathcal{L}(x_0(\hat{\theta}), y_{tar}, x_0, y_{ref}))$}\\
        \BlankLine
        \textcolor{dartmouthgreen}{
        $x_0(t_0; \hat{\theta}) = \left(x_{t_0} - \sqrt{1 - \alpha_{t_0}}\epsilon_{\theta}(x_{t_0}, t_0)\right) / {\sqrt{\alpha_{\tau}}}$
        \BlankLine
        $\hat{\theta} = \hat{\theta} + \nabla_{\hat{\theta}} \mathcal{L}(x_0(t_0; \hat{\theta}), y_{tar}, x_0, y_{ref}))$
        }
    }
    \Return $\epsilon_{\hat{\theta}}$
\label{alg:training}
\end{algorithm}

\section{Technical setup}
\label{appendix:tech_setup}
All experiments are performed on a single Tesla A100 GPU. 
64 CPU cores are used. PyTorch version is 1.10.1.

We use the following architectures of the diffusion models: \cite{ho2020ddpm} for Celeba-HQ-256, \cite{pmlr-v139-nichol21a} for LSUN-church-256, AFHQ-dog-256 and ImageNet-512.
\label{appendix:setup}

%%%%%%%%%%%%%%%%%%%%%%%%%%%%%%%%%%%%%%%%%%%%%%%%%%%%%%%%%
\section{Evaluation of stochastic encoding}
\label{appendix:mos_encoding}
In this section, we compare stochastic (DDPM) and deterministic (DDIM) encoding methods using the mean opinion score. We ask assessors to estimate how the reconstructed image is similar to the reference one according to the criteria in \tab{mos_encoding} (Bottom).

According to the human opinion in \tab{mos_encoding} (Top), for $t_{0} = 300$, DDPM encoding does not affect semantic attributes noticeably. Thus, our method considers lower $t_0\le300$ for shallow image manipulations, e.g., ``Makeup face'', where preserving most details of the original image is important.
On the other hand, $t_{0} = 500$ slightly alters the face attributes. Therefore, we use $t_0{=}350{-}500$ for severe transforms, e.g., ``Zombie'', that significantly affect the semantic attributes.
%%%%%%%%%%%%%%%%%%%%%%%%%%%%%%%%%%%%%%%%%%%%%%%%%%%%%%%%%%

%%%%%%%%%%%%%%%%%%%%%%%%%%%%%%%%%%%%%%%%%%%%%%%%%%%%%
\begin{table}
\centering
\resizebox{4.0cm}{!}{
 \begin{tabular}{|c| c c c |} 
 \hline
 $t_{0}$ & 100 & 300 & 500 \\ 
 \hline
 DDPM & $4.58$ & $4.08$ & $3.40$\\   
 DDIM & $4.75$ & $4.46$ & $4.40$\\ 
 \hline
 \end{tabular}} \\
 \vspace{2mm}
 \resizebox{8.2cm}{!}{
 \begin{tabular}{|c|} 
 \hline
 Criterion \\ 
 \hline
5: Looks identical. No visible changes or artifacts. \\   
4: Minor visible changes. All face attributes are fully preserved. \\ 
3: Minor changes of the face attributes. The person is the same. \\
2: Significant changes of the face attributes.\\
1: Completely different person. \\ 
 \hline
 \end{tabular}}
 \caption{\textbf{(Top)} Comparison of stochastic (DDPM) and deterministic (DDIM) encoding methods for different steps $t_{0}$ in terms of mean opinion score that measures similarity between reconstructed and original images. \textbf{(Bottom)} The criteria which were shown to assessors for estimation.}
\label{tab:mos_encoding}
\end{table}

\section{List of textual transforms}
\subsection{Prelearned image manipulations}
\label{appendix:domain_transform_list}
For the human evaluation, we use the following established transformations \cite{kim2022diffusionclip, gal2021stylegannada}:
\begin{itemize}%[leftmargin=10px]
    \item \textbf{Celeba-HQ-256} ---  ``Face'' $\rightarrow$ ``Angry face'', ``Face'' $\rightarrow$ ``Pale face'', ``Face'' $\rightarrow$ ``Smiling Face'', ``Photo'' $\rightarrow$ ``Painting in Fernando Botero style'', ``Person'' $\rightarrow$ ``Nicolas Cage'', ``Photo'' $\rightarrow$ ``Painting in Cubism style'', ``Face'' $\rightarrow$ ``Makeup face'', ``Photo'' $\rightarrow$ ``Painting in Modigliani style'', ``Human'' $\rightarrow$ ``Neanderthal, ``Person'' $\rightarrow$ ``Old person'', ``Human'' $\rightarrow$ ``3D rendering in style of Pixar'', ``Person'' $\rightarrow$ ``Surprised'', ``Face'' $\rightarrow$ ``Tanned face'', ``Photo'' $\rightarrow$ ``Watercolor art'', ``Human'' $\rightarrow$ ``Zombie'', ``Person'' $\rightarrow$ ``Mark Zuckerberg''.
    \item \textbf{LSUN-church-256} ---  ``Church'' $\rightarrow$ ``Golden church, Church'' $\rightarrow$ ``Colorful church'', ``Church'' $\rightarrow$ ``Gothic church, ``Church'' $\rightarrow$ ``Modern architecture'', ``Church'' $\rightarrow$ ``Snow covered church'', ``Church'' $\rightarrow$ ``Ancient temple''.
    \item \textbf{AFHQ-dog-256} ---  ``Dog'' $\rightarrow$ ``Angry dog'', ``Dog'' $\rightarrow$ ``Anime dog'', ``Dog'' $\rightarrow$ ``Bear'', ``Dog'' $\rightarrow$ ``Fox'', ``Dog'' $\rightarrow$ ``Smiling dog'', ``Dog to Zombie dog''.
    \item \textbf{ImageNet-512} --- ``Photo'' $\rightarrow$ ``Painting in cubism style'', ``Photo'' $\rightarrow$ ``Painting in cubism style'', ``Photo'' $\rightarrow$ ``Painting in Van Gogh style'', ``Photo'' $\rightarrow$ ``Painting in pointilism style'', ``Photo'' $\rightarrow$ ``Sketch'', ``Photo'' $\rightarrow$ ``Watercolor art''. 
\end{itemize}

\subsection{Single image editing}
\label{appendix:sin_transform_list}
\begin{itemize}%[leftmargin=10px]
    \item \textbf{In-domain} transforms --- ``Face'' $\rightarrow$ ``Angry face'', ``Person'' $\rightarrow$ ``Nicolas Cage'', ``Face'' $\rightarrow$ ``Makeup face'', ``Person'' $\rightarrow$ ``Old person'', ``Person'' $\rightarrow$ ``Surprised'', ``Man'' $\rightarrow$ ``Woman'' (or vice versa), ``Face'' $\rightarrow$ ``Smiling face'', ``Person'' $\rightarrow$ ``Mark Zuckerberg''.
    \item \textbf{Out-of-domain} transforms --- ``Human'' $\rightarrow$ ``3D rendering in style of Pixar'', ``Human'' $\rightarrow$ ``Neanderthal'', ``Photo'' $\rightarrow$ ``Painting in Fernando Botero style'', ``Photo'' $\rightarrow$ ``Painting in Modigliani style'', ``Photo'' $\rightarrow$ ``Self portrait by Frida Kahlo'', ``Photo'' $\rightarrow$ ``Sketch'', ``Human'' $\rightarrow$ ``Jocker'', ``Human'' $\rightarrow$ ``Zombie''. 
\end{itemize}

\begin{table}[h!]
\centering
\resizebox{7.0cm}{!}{
 \begin{tabular}{|c| c c c|} 
 \hline
 Text description & $lr$ & $\lambda$ & $t_{0}$\\ [0.5ex] 
 \hline\hline
 &\multicolumn{3}{c|}{\textbf{CelebA-HQ-256}} \\ [1ex] 
  \hline
 Angry face & $3e{-}6$ & 0.6 & 300 \\  
 Pale face & $3e{-}6$ & 0.6 & 300 \\  
 Smiling face & $3e{-}6$ & 0.6 & 300 \\  
 Painting in Fernando Botero style & $8e{-}6$ & 0.3 & 450 \\ 
 Nicolas Cage & $7e{-}6$ & 0.1 & 450 \\ 
 Painting in Cubism style & $7e{-}6$ & 0.1 & 450 \\ 
 Makeup face & $1e{-}6$ & 0.9 & 300 \\ 
 Painting in Modigliani style & $5e{-}6$ & 0.4 & 450 \\ 
 Neanderthal & $6e{-}6$ & 0.3 & 450 \\
 Old person & $5e{-}6$ & 0.2 & 380 \\ 
 3D rendering in style of Pixar & $3e{-}6$ & 0.1 & 450 \\
 Surprised & $4e{-}6$ & 0.8 & 450 \\
 Tanned face & $4e{-}6$ & 0.8 & 450 \\ 
 Watercolor art & $6e{-}6$ & 0.5 & 450 \\
 Zombie & $2e{-}5$ & 0.1 & 430 \\ 
 Mark Zuckerberg & $2e{-}6$ & 0.3 & 400 \\ [1ex] 
 \hline \hline
 &\multicolumn{3}{c|}{\textbf{LSUN-Church-256}} \\ [1ex] 
 \hline
  Golden church & $3e{-}6$ & 0.1 & 380 \\ 
  Colorful church & $3e{-}6$ & 0.0 & 400 \\  
  Gothic church & $3e{-}6$ & 0.1 & 380 \\  
  Modern architecture & $2e{-}6$ & 0.2 & 400\\
  Snow covered church & $3e{-}6$ & 0.6 & 400 \\ Ancient temple & $4e{-}6$ & 0.8 & 400 \\ [1ex]
 \hline
  \hline 
 &\multicolumn{3}{c|}{\textbf{AFHQ-Dog-256}} \\ [1ex] 
 \hline
  Angry dog & $6e{-}6$ & 0.8 & 380 \\ 
  Anime dog & $8e{-}6$ & 0.2 & 450 \\  
  Bear & $8e{-}6$ & 0.7 & 450 \\  
  Fox & $8e{-}6$ & 0.6 & 450\\
  Smiling dog & $6e{-}6$ & 0.3 & 400 \\ 
  Zombie dog& $6e{-}6$ & 0.2 & 450 \\ [1ex]
 \hline
   \hline
 &\multicolumn{3}{c|}{\textbf{ImageNet-512}} \\ [1ex] 
 \hline
  Painting in cubism style & $8e{-}6$ & 0.1 & 450 \\ 
  Painting in cubism style & $8e{-}6$ & 0.1 & 450 \\  
  Painting in Van Gogh style & $8e{-}6$ & 0.1 & 450 \\  
  Painting in pointilism style & $8e{-}6$ & 0.2 & 450\\
  Sketch & $8e{-}6$ & 0.1 & 450 \\ 
  Watercolor art& $6e{-}6$ & 0.1 & 450 \\ [1ex]
 \hline
 \end{tabular}}
 \caption{Hyperparameter values used to learn text-driven image manipulations. $\lambda$ represents the coefficient in front of the $\mathcal{L}_{id}$ term.}
 \label{tab:hyperparameters}
\end{table}

\begin{figure}[H]
    \centering
    \includegraphics[width=6.cm]{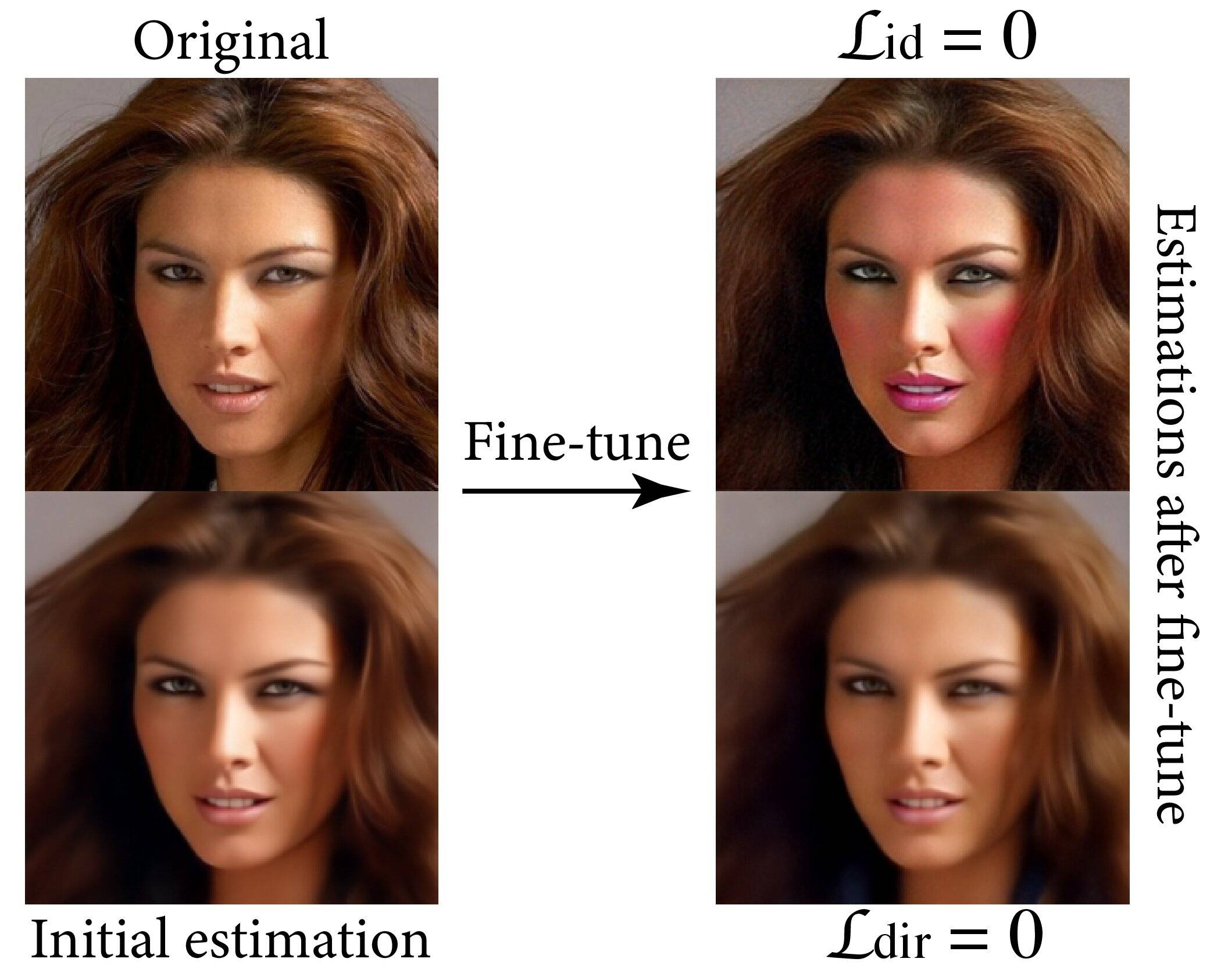}
    \caption{The $x_0$ estimates before and after finetuning using $\mathcal{L}_{dir}$ and $\mathcal{L}_{id}$ losses, independently. Opposed to $\mathcal{L}_{dir}$, $\mathcal{L}_{id}$ does not improve the perceptual quality of the estimates.}
    \label{fig:111}
\end{figure}

\section{Quality improvement effect}
\label{appendix:quality_imp}
The paper demonstrates that our adaptation procedure improves the perceptual quality  of the approximate $x_0$ estimates at a time step $t_0$. 
% First, we present more visual examples of the $x_0$ estimates before and after finetuning in \fig{quality_effect_imp}.

In this experiment, we ensure that $\mathcal{L}_{dir}$~\eq{clip_loss} is indeed responsible for these changes but not $\mathcal{L}_{id}$~\eq{objective}.
\fig{111} provides the visual examples of the $x_0$ estimates after the model finetuning independently using one of these losses. 
We observe that $\mathcal{L}_{id}$ does not lead to perceptual quality improvements.

%%%%%%%%%%%%%%%%%%%%%%%%%%%
%%%%%%%%%%%%%%%%%%%%%%%%%%%
\begin{figure}[H]
    \centering
    \includegraphics[width=7cm]{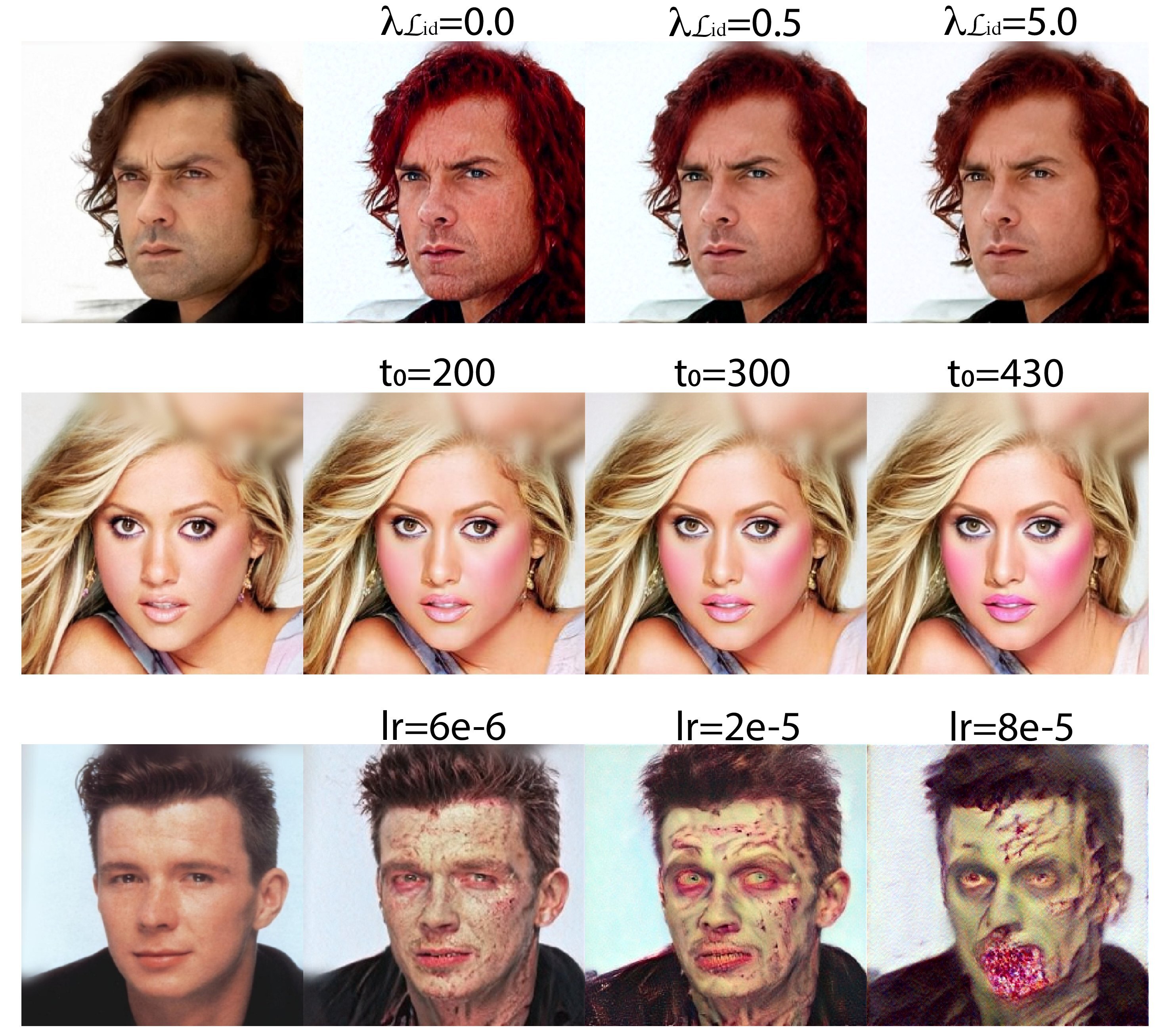}
    \caption{The effect of different hyperparameter values for a few transforms. $\lambda_{\mathcal{L}_{id}}$ corresponds to the coefficient in front of the regularization term.}
    \label{fig:hyper}
\end{figure}

\section{Hyperparameters}
\label{appendix:hyper}

The proposed approach is sensitive to hyperparameter values. 
In \tab{hyperparameters}, we provide the full list of hyperparameter values used for different datasets and transforms.
\fig{hyper} presents a few examples of image manipulations with different hyperparameter values. 
Below, we give a few recommendations to make the tuning of our method more approachable for users.

\begin{itemize}
    \item Consider using higher values of the regularization coefficient $\lambda$ (e.g., $5{-}10$) for the shallow transforms like ``Red hair''.
    \item Higher values of the $t_{0}$ allow us to make stronger transforms but can bring more irrelevant changes. 
    Thus, we recommend to use smaller $t_{0}$ values ($t_{0} = 200{-}350$) for the shallow transforms, e.g., ``Makeup face'', ``Surprised face'', and higher $t_{0}$ values ($t_{0} = 350{-}500$) for the strong manipulations, e.g., ``Zombie'', ``Sketch''.
    \item We also find that a higher learning rate, e.g., $2e{-}5$, allows obtaining better results for the strong transforms, while for the shallow ones, smaller values are preferable, e.g., $5e{-}6$.
\end{itemize}
%%%%%%%%%%%%%%%%%%%%%%%%
%%%%%%%%%%%%%%%%%%%%%%%%
\section{Failure cases}

\begin{figure}
    \centering
    \includegraphics[width=7.9cm]{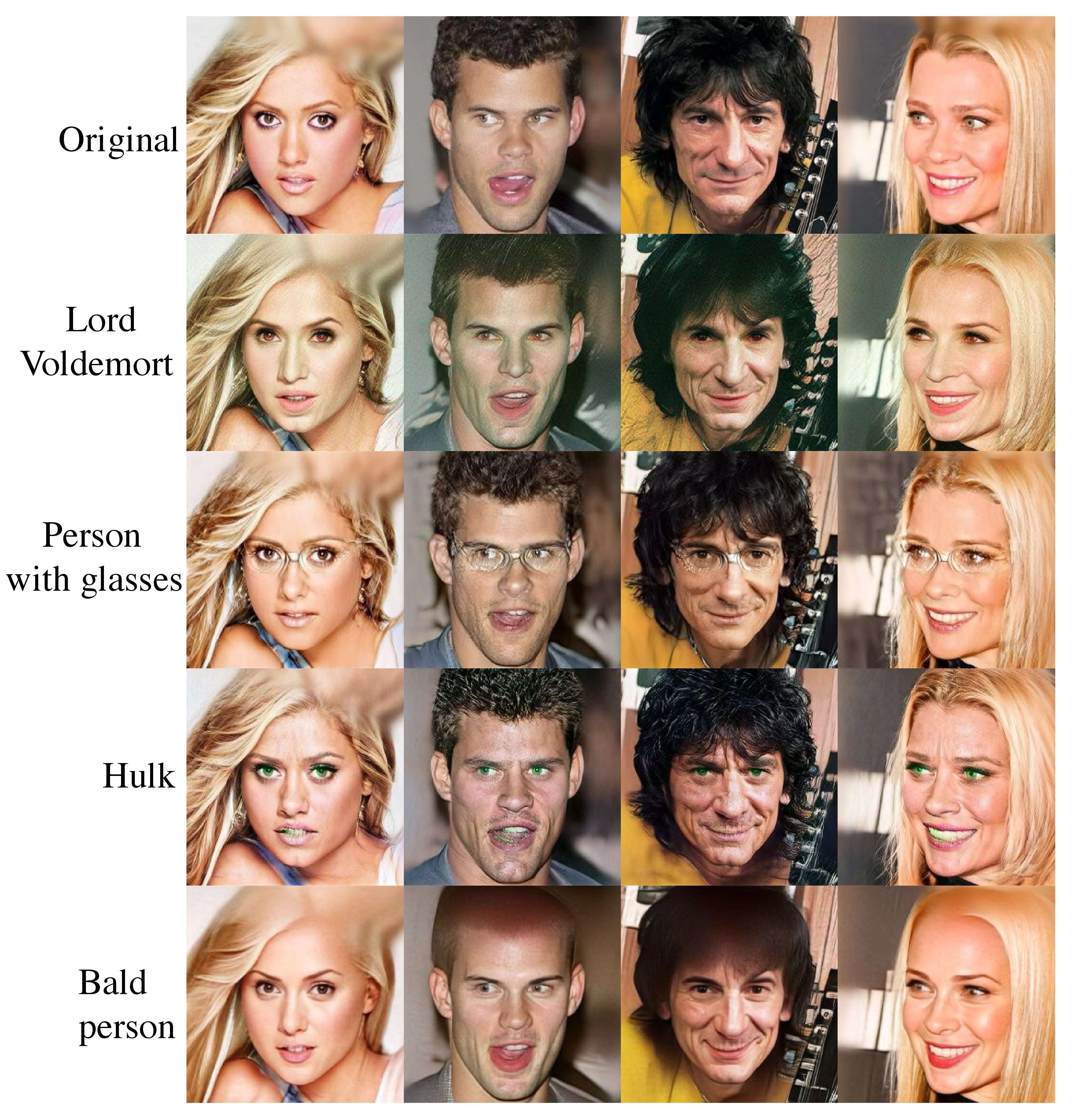}
    \caption{Visual examples of the failed image manipulations produced with our approach.}
    \label{fig:fails}
\end{figure}

\begin{figure}
    \centering
    \includegraphics[width=8.5cm]{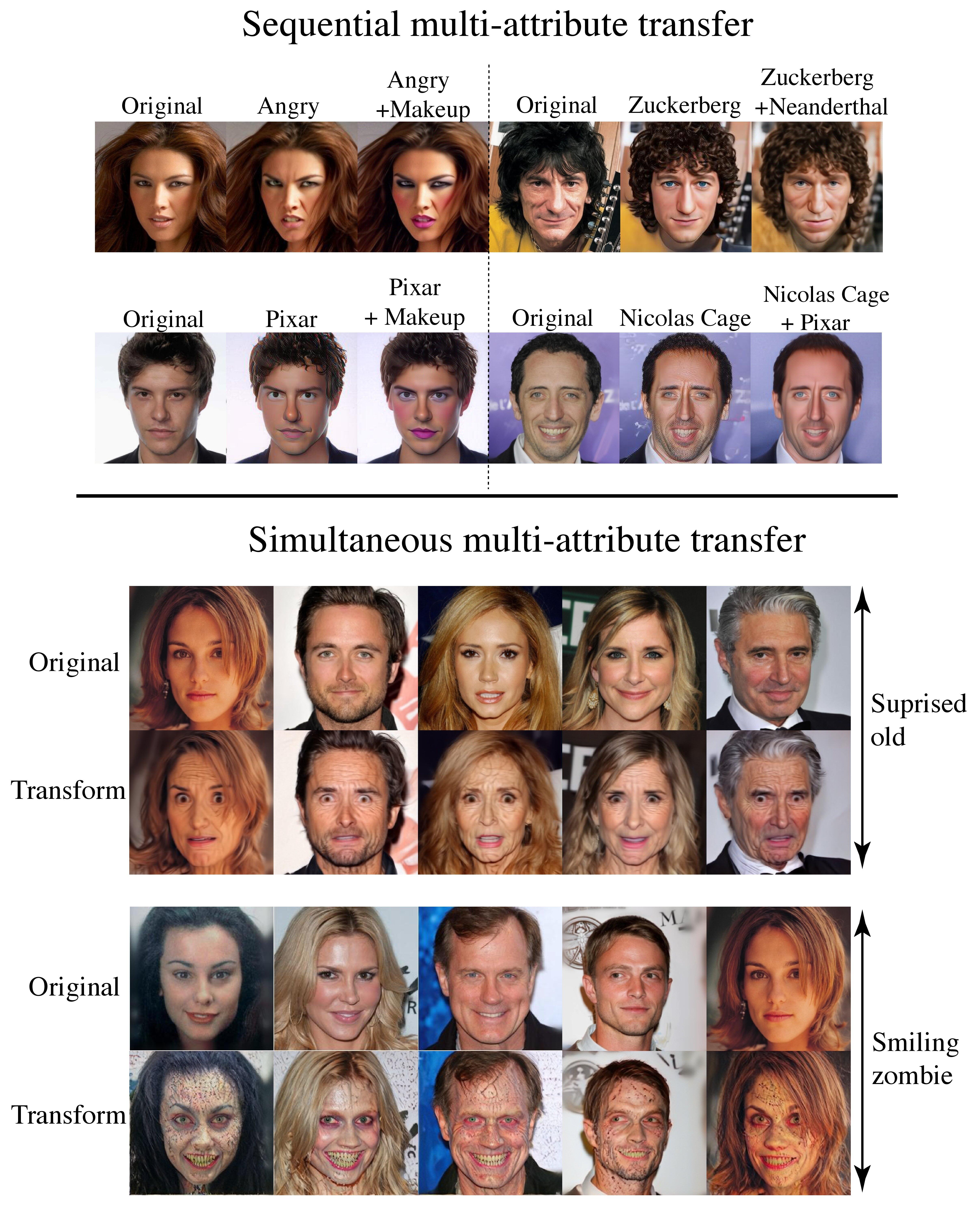}
    \caption{Visual examples of the multi-attribute transfer. \textbf{Sequential} setting adapts the model to two different text descriptions one by one.
    \textbf{Simultaneous} means that we form a single text description that combines two different attributes.}
    \label{fig:combo}
\end{figure}

The proposed approach is limited by the capabilities of the CLIP and pretrained diffusion models. 
Sometimes, the former cannot provide the desired signal to adapt the diffusion model properly. 
In \fig{fails}, we provide a few visual examples of such text descriptions.
%%%%%%%%%%%%%%%%%%%%%%%%

%%%%%%%%%%%%%%%%%%%%%%%%
\section{Multi-attribute transfer}

Following~\cite{kim2022diffusionclip}, we also demonstrate that our approach is able to produce multi-attribute transfers in both sequential and simultaneous regimes. 
The visual examples of the learned image manipulations are presented in \fig{combo}.

\begin{figure*}
    \centering
    \includegraphics[width=19.5cm]{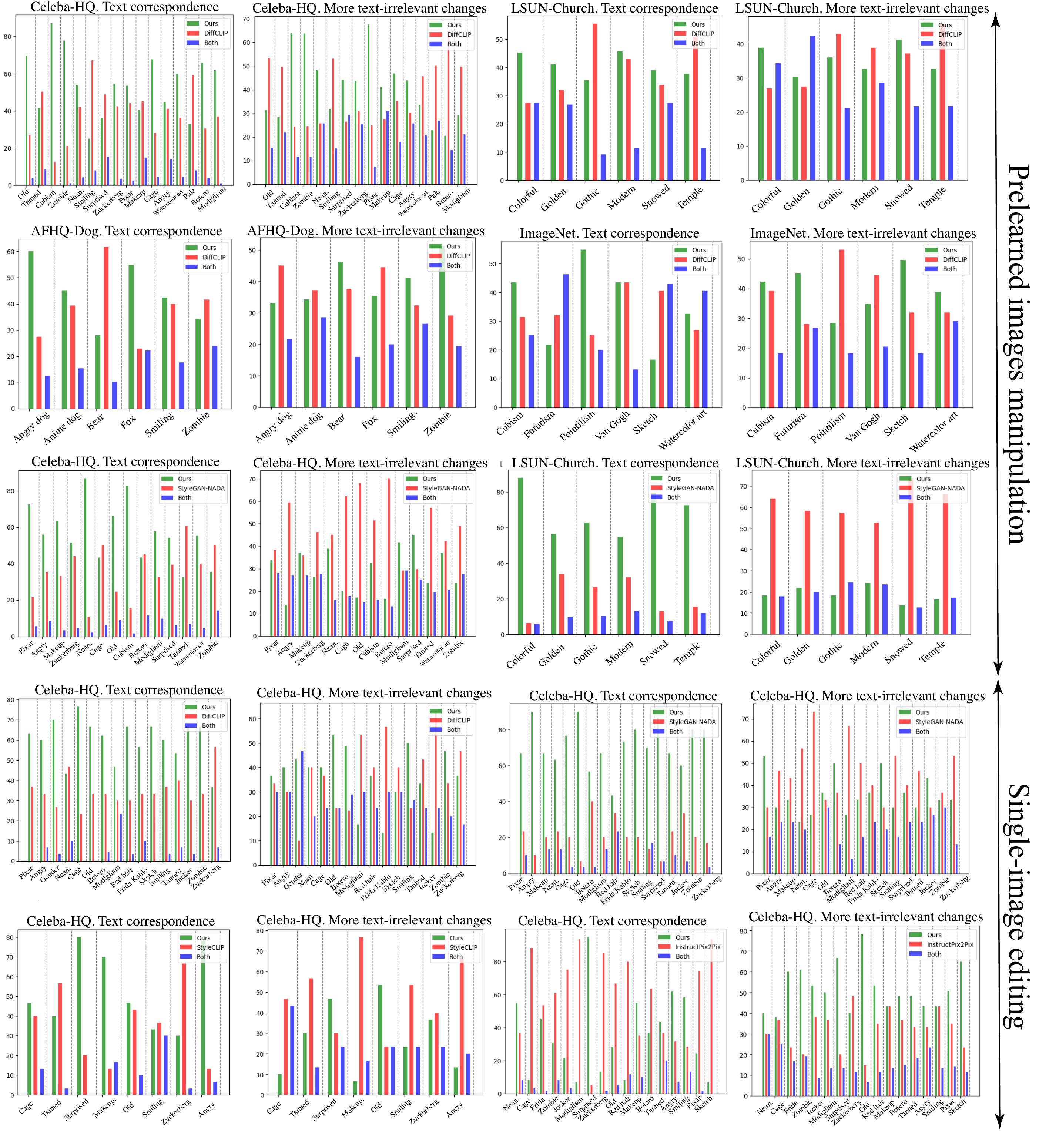}
    \caption{Voting results for prelearned image manipulations (Top) and single-image editing (Bottom).}
    \label{fig:diag}
\end{figure*}

\begin{figure*}
    \centering
    \includegraphics[width=13.5cm]{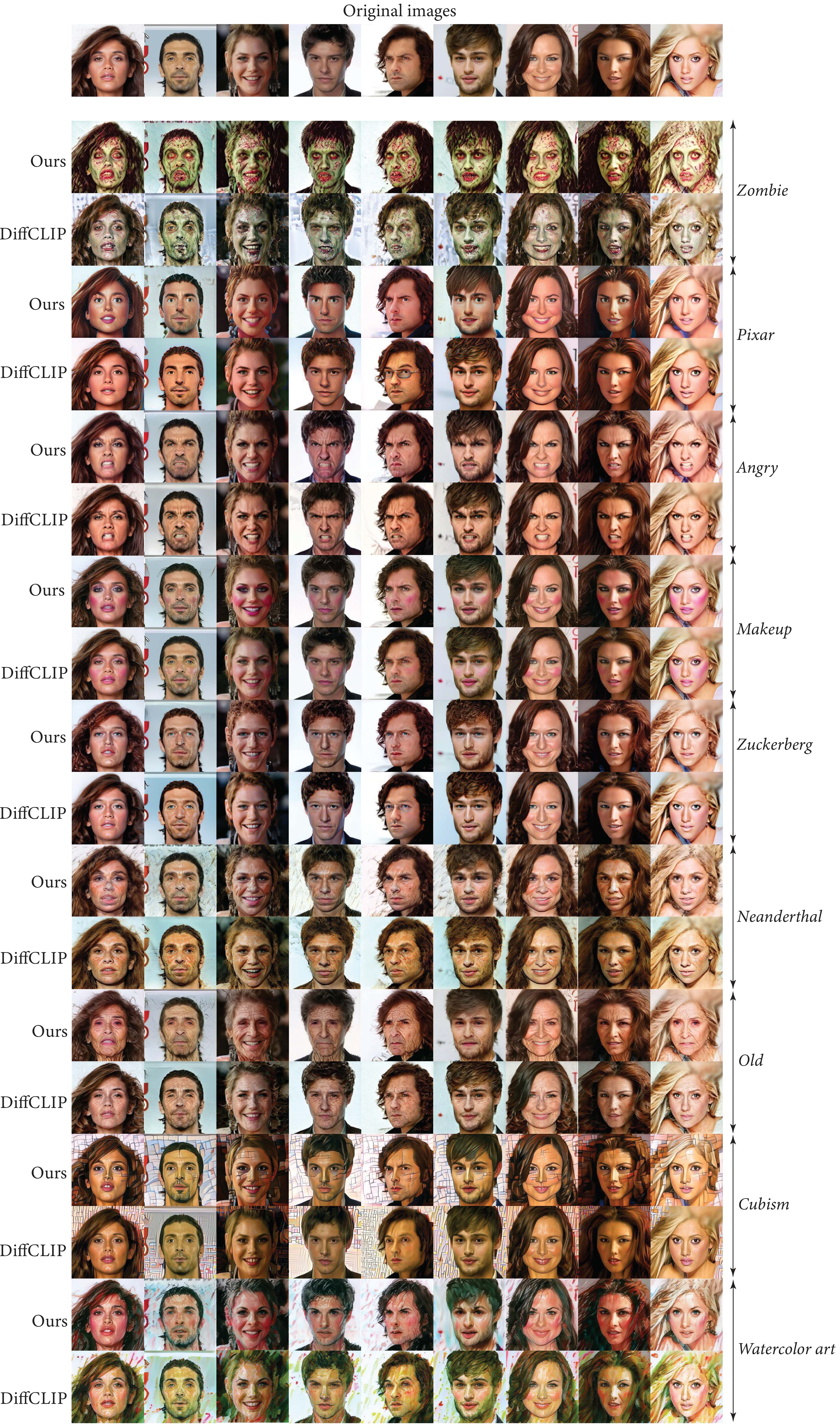}
    \caption{More visual examples of the prelearned image manipulations on CelebA-HQ-256 (Ours and DiffusionCLIP).}
    \label{fig:comparison_adapt}
\end{figure*}

\begin{figure*}
    \centering
    \includegraphics[width=15.5cm]{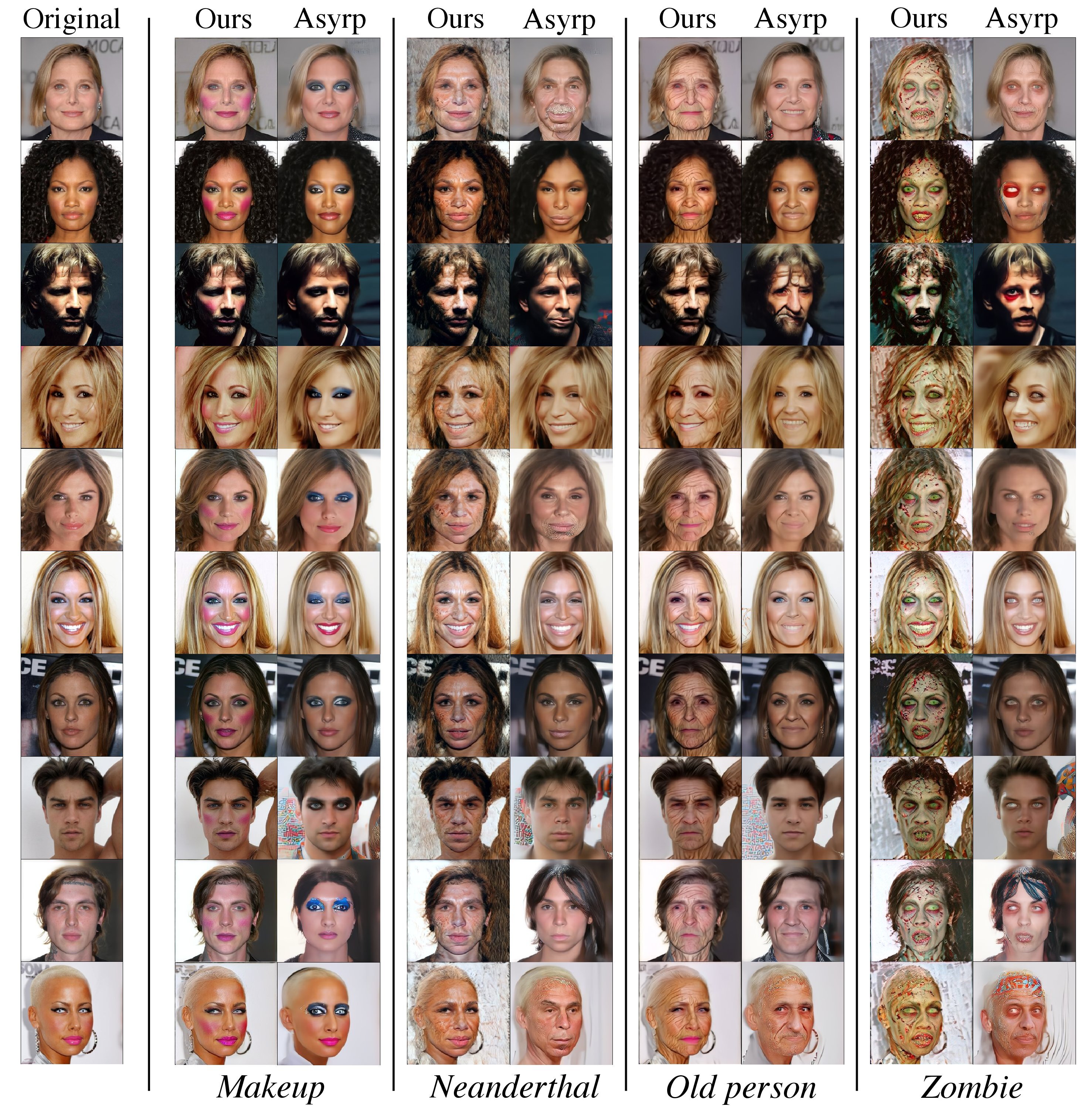}
    \caption{More visual examples of the prelearned image manipulations on CelebA-HQ-256 (Ours and Asyrp).}
    \label{fig:comparison_semantic}
\end{figure*}

\begin{figure*}
    \centering
    \includegraphics[width=13.5cm]{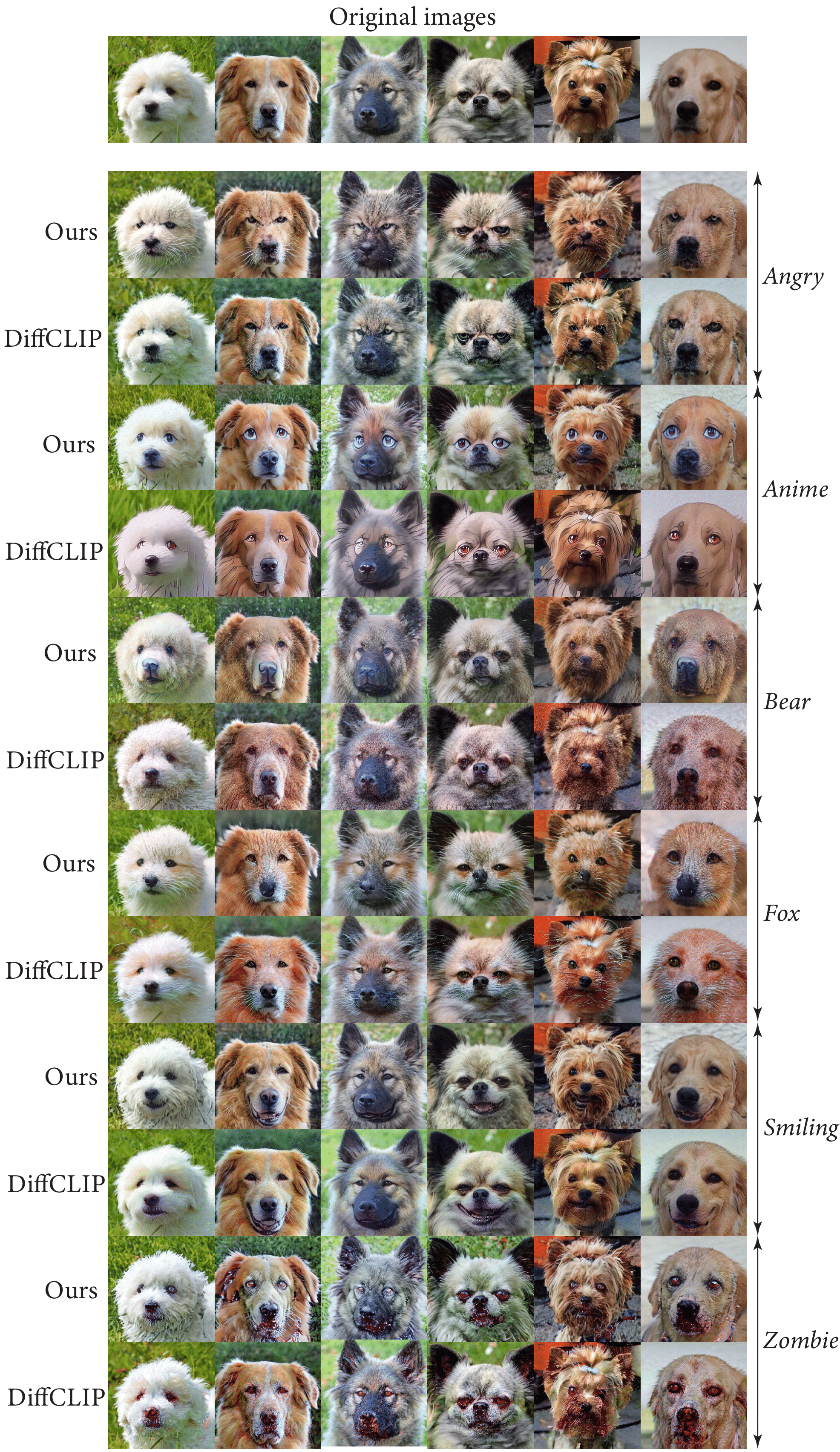}
    \caption{More visual examples of the prelearned image manipulations on AFHQ-Dogs-256 (Ours and DiffusionCLIP).}
    \label{fig:comparison_dogs}
\end{figure*}

\begin{figure*}
    \centering
    \includegraphics[width=16cm]{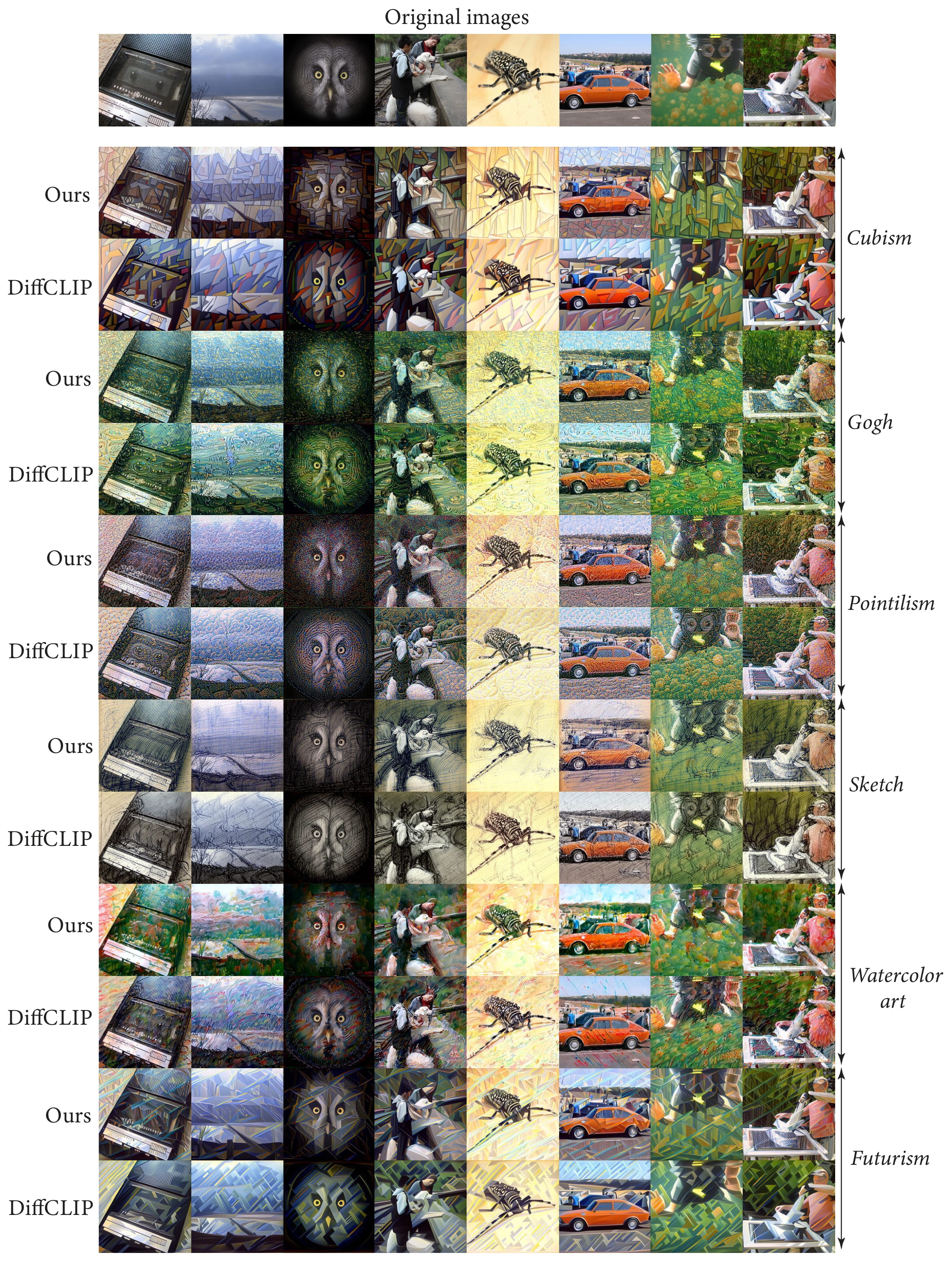}
    \caption{More visual examples of the prelearned image manipulations on ImageNet-512 (Ours and DiffusionCLIP).}
    \label{fig:comparison_imagenet}
\end{figure*}

\begin{figure*}
    \centering
    \includegraphics[width=16cm]{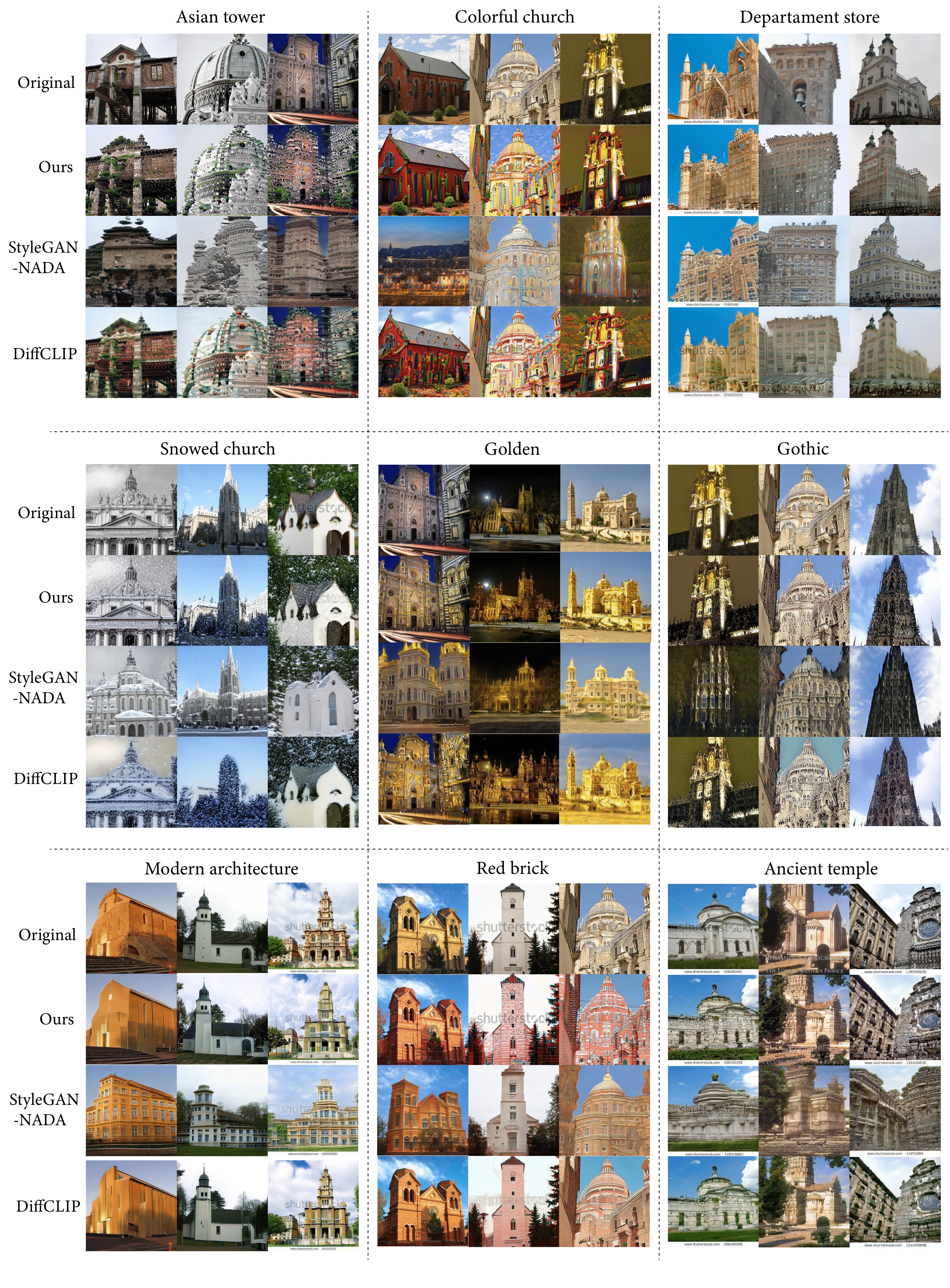}
    \caption{More visual examples of the prelearned image manipulations on LSUN-Church-256 (Ours, DiffusionCLIP and StyleGAN-NADA).}
    \label{fig:comparison_church}
\end{figure*}

\begin{figure*}
    \centering
    \includegraphics[width=16cm]{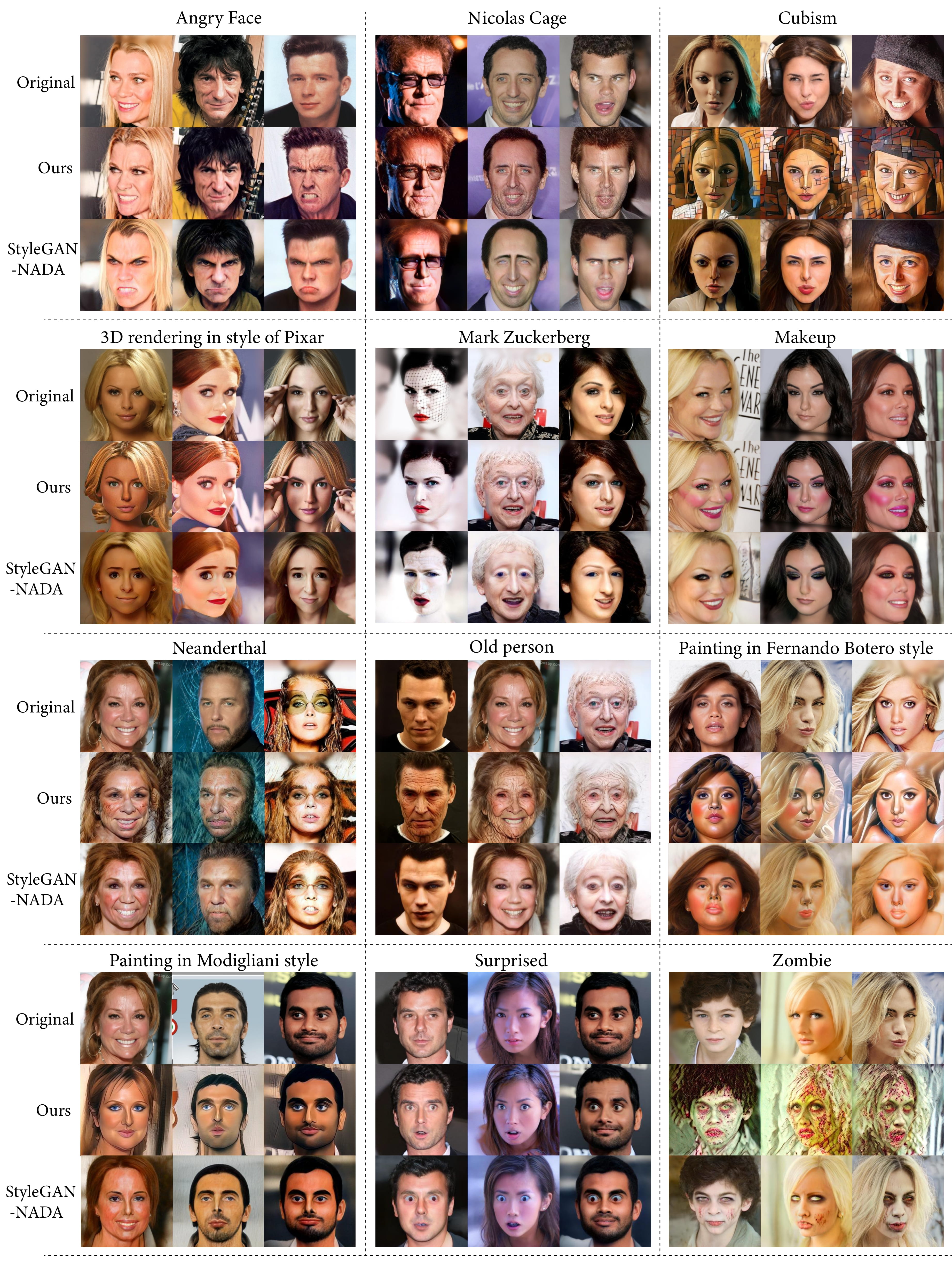}
    \caption{More visual examples of the prelearned image manipulations on CelebA-HQ-256 (Ours and StyleGAN-NADA).}
    \label{fig:comparison_gans}
\end{figure*}

\begin{figure*}
    \centering
    \includegraphics[width=13.5cm]{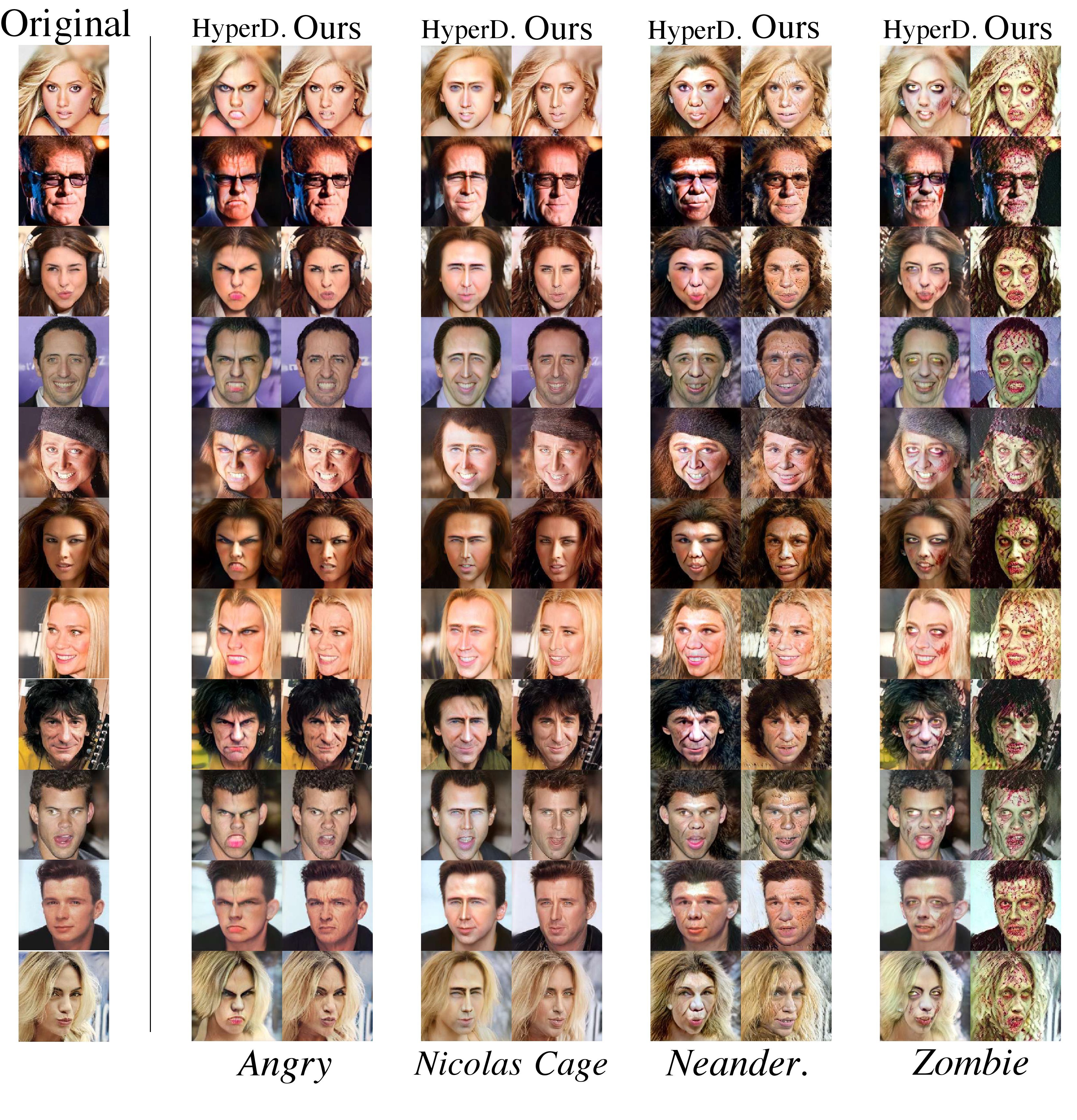}
    \caption{More visual examples of the prelearned image manipulations on CelebA-HQ-256 (Ours and HyperDomainNet).}
    \label{fig:comparison_hyper}
\end{figure*}

\begin{figure*}
    \centering
    \includegraphics[width=9.6cm]{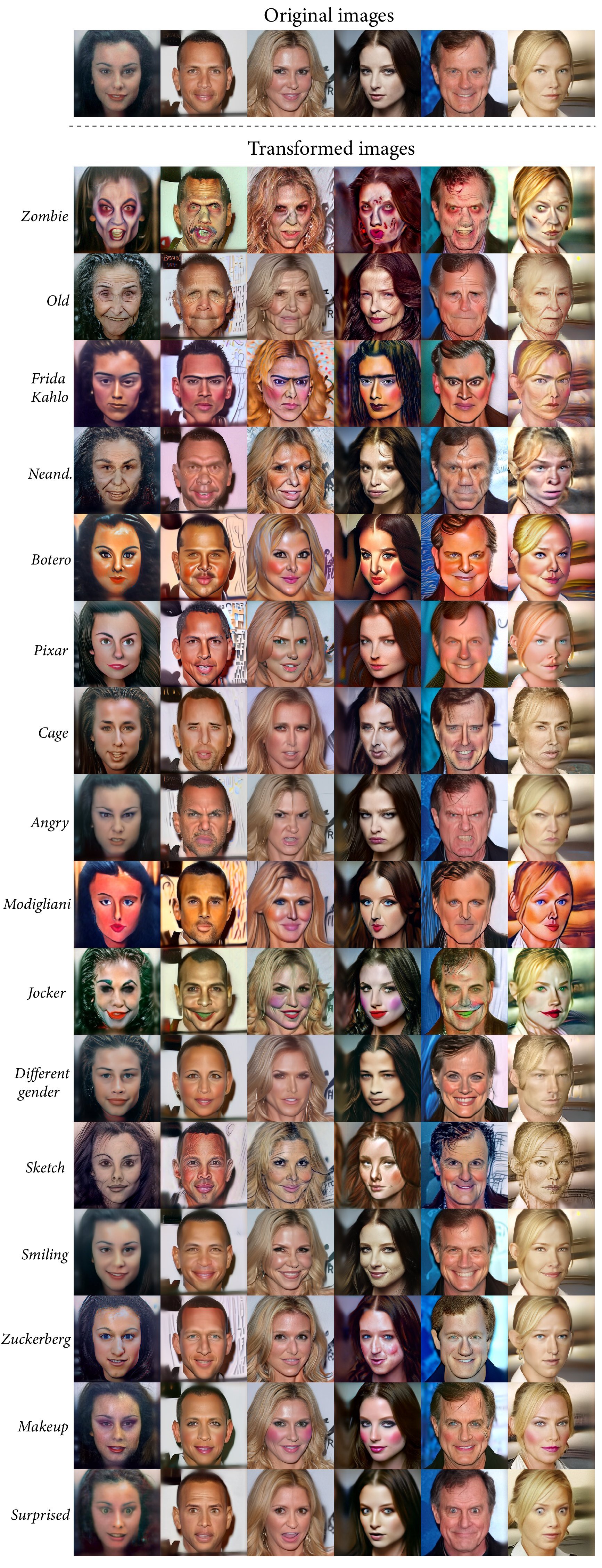}
    \vspace{-3mm}
    \caption{More visual examples of the text-guided single-image editing produced with our method.}
    \label{fig:sin_images}
\end{figure*}

\begin{figure*}
    \centering
    \includegraphics[width=12cm]{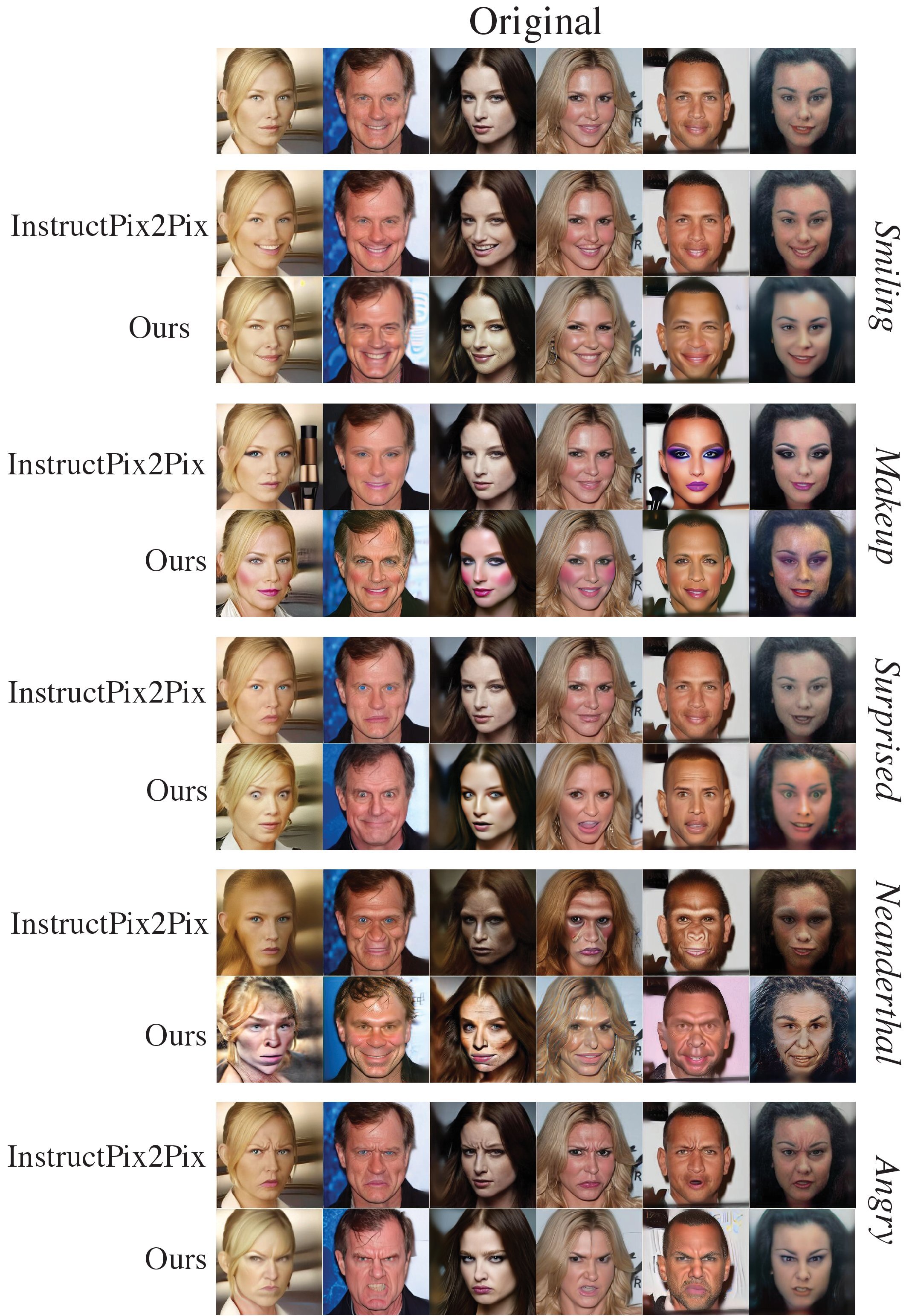}
    \caption{More visual examples of the text-guided single-image editing (Ours and InstructPix2Pix).}
    \label{fig:sin_pix2pix}
\end{figure*}

\end{document}